\documentclass[sigconf]{acmart}

\renewcommand\footnotetextcopyrightpermission[1]{} 

\usepackage{multirow} 
 
\usepackage{amsthm,amsmath,amssymb}
\usepackage{mathrsfs}
\usepackage{graphicx}
\usepackage{float}
\usepackage{subfigure}
\usepackage{hyperref}
\hypersetup{hidelinks,
	colorlinks=true,
	allcolors=black,
	pdfstartview=Fit,
	breaklinks=true}
 
\AtBeginDocument{%
  }

\acmConference[Conference acronym 'XX]{Make sure to enter the correct
  conference title from your rights confirmation emai}{June 03--05,
  2023}{Woodstock, NY}
\acmPrice{15.00}
\acmISBN{978-1-4503-XXXX-X/18/06}

\acmSubmissionID{2080}

\begin{document}

\title{StylePrompter: All Styles Need Is Attention}

\author{Chenyi Zhuang}
\affiliation{%
  \institution{Nanjing University of Aeronautics and Astronautics}
  \city{Nanjing}
  \country{China}
}
\email{chenyi.zhuang@nuaa.edu.cn}

\author{Pan Gao}
\authornote{Corresponding author}
\affiliation{%
  \institution{Nanjing University of Aeronautics and Astronautics}
  \city{Nanjing}
  \country{China}
}
\email{pan.gao@nuaa.edu.cn}

\author{Aljosa Smolic}
\affiliation{%
  \institution{Lucerne University of Applied Sciences and Arts}
  \city{Lucerne}
  \country{Switzerland}
}
\email{aljosa.smolic@hslu.ch}

\pagestyle{plain} 
\renewcommand{\shortauthors}{Zhuang et al.}

\begin{abstract}
GAN inversion aims at inverting given images into corresponding latent codes for Generative Adversarial Networks (GANs), especially StyleGAN where exists a disentangled latent space that allows attribute-based image manipulation at latent level. As most inversion methods build upon Convolutional Neural Networks (CNNs), we transfer a hierarchical vision Transformer backbone innovatively to predict $\mathcal{W^+}$ latent codes at token level. We further apply a Style-driven Multi-scale Adaptive Refinement Transformer (SMART) in $\mathcal{F}$ space to refine the intermediate style features of the generator. By treating style features as queries to retrieve lost identity information from the encoder's feature maps, SMART can not only produce high-quality inverted images but also surprisingly adapt to editing tasks. We then prove that StylePrompter lies in a more disentangled $\mathcal{W^+}$ and show the controllability of SMART. Finally, quantitative and qualitative experiments demonstrate that StylePrompter can achieve desirable performance in balancing reconstruction quality and editability, and is "smart" enough to fit into most edits, outperforming other $\mathcal{F}$-involved inversion methods.
Our code is available at: \href{https://github.com/I2-Multimedia-Lab/StylePrompter}{https://github.com/I2-Multimedia-Lab/StylePrompter}.
\end{abstract}



\maketitle

\section{Introduction}

StyleGAN and its family \cite{karras2019style, karras2020analyzing, karras2020training, karras2021alias} emerge victorious in Generative Adversarial Networks (GANs) not only for high-quality generated images but also an intermediate latent space with disentangled attributes, making latent-based image manipulation possible \cite{richardson2021encoding}. However, only randomly generated images can enjoy this editability. The lack of making inferences on a target image in generic GANs gives birth to GAN inversion, which can map any given image into latent codes in specific StyleGAN latent space for better application.

\begin{figure}[h]
  \centering
  \includegraphics[scale=0.64]{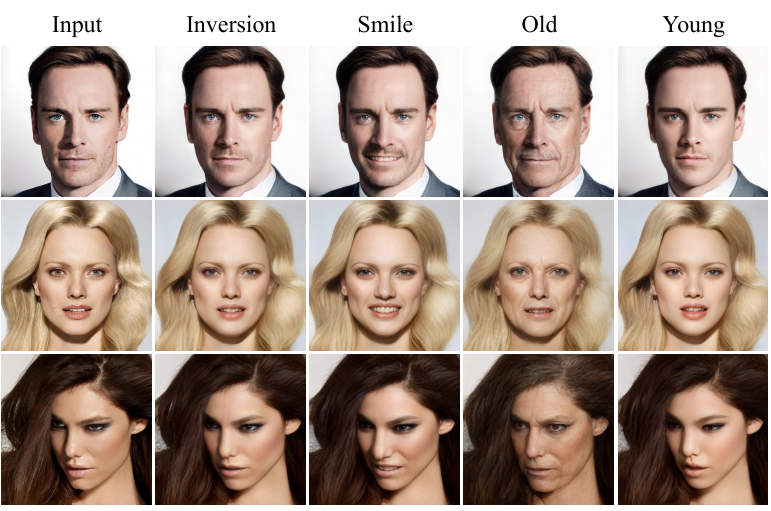}
  \caption{StylePrompter with SMART can fulfill high-quality inversion results and maintain the editability of $\mathcal{F}$.}
  \vspace{-0pt}
  \Description{headfigure}
  \label{headfigure}
\end{figure}

In StyleGAN, there exists several latent spaces, denoted as $\mathcal{Z}$,$\mathcal{W}$, $\mathcal{W^+}$, $\mathcal{S}$, $\mathcal{F}$. 
To be specific, $\mathcal{Z}$ is the original space where generative models learn to map from, usually a normal or uniform distribution. 
Then a mapping network converts $z \in \mathcal{Z}$ into a more disentangled latent space $\mathcal{W}$, which does not follow any distribution \cite{karras2019style, tov2021designing}. 
The Synthesis network of StyleGAN stacked by convolution layers will progressively increase the image resolution, where the convolution kernel weights are channel-wise style codes $s \in \mathcal{S}$, specialized by the latent codes $w \in \mathcal{W}$.
$\mathcal{W^+}$ comes out as an extended space for $\mathcal{W}$ that allows different $w$ inputs for convolution layers \cite{abdal2019image2stylegan, abdal2020image2stylegan++, richardson2021encoding}. Finally, all output tensors of convolution layers define a feature space $\mathcal{F}$. 
We found this synthesis of StyleGAN can be seen as oil paintings. Latent codes in $\mathcal{W}$ space are \textit{pigments}, controlling \textbf{WHICH} attributes to generate. $\mathcal{S}$ space is like \textit{brush}, deciding \textbf{HOW} to add these attributes, and the deepest $\mathcal{F}$ space is the \textit{canvas} to show \textbf{WHAT} is drawn.
This vivid example can also fit the theory proven by previous works: despite the Gaussian-like $\mathcal{Z}$ space which is not disentangled enough for editing task, the expression of latent codes sampled from $\mathcal{W}$, $\mathcal{W^+}$, $\mathcal{S}$ and $\mathcal{F}$ space increase in sequence, however, their editability show an opposite trend, i.e., deeper space would be more difficult to control \cite{yao2022feature, kang2021gan}.

The main concern of GAN inversion is to obtain latent codes corresponding to high-quality inverted images and can be edited flexibly at the same time, which is not easy since the trade-off aforesaid, formulated as \textit{distortion-editability} in previous works \cite{tov2021designing, hu2022style, roich2022pivotal}. Whereas the term \textit{distortion} is not expressive enough to reflect aesthetic perception, we instead use \textit{quality} as a combination of fidelity and realism.
Particularly, fidelity estimates the similarity between the input image and the inverted one, yet realism is highly related to the perceptual preference in the human vision system that the inverted image should not have unreal textures or artifacts. $\mathcal{F}$ with the widest manifold is the best choice for high-quality inversion, but is the most difficult space to manipulate, showing a dilemma of \textit{quality-editability}. Many works have made efforts to address this issue, most however through CNN-based architectures.  It remains to explore whether other architectures like Transformers are effective for this trade-off or not.

In this paper, we adopt a hierarchical vision Transformer to tackle this quality-editability trade-off, build a novel lightweight Transformer-based framework, \textbf{StylePrompter}, as latent codes are embedded as tokens, similar to \textit{prompts}, interacting with patch tokens of the image; and we treat style features as \textit{prompter} to retrieve the lost identity information in our proposed Style-driven Multi-scale Adaptive Refinement Transformer (SMART) block. Our contributions can be summarized as follows:
\begin{itemize}
\item We propose a novel Transformer-based backbone to predict $\mathcal{W^+}$ latent codes at token level. To our knowledge, we are the first to adopt such a hierarchical vision Transformer backbone for GAN inversion.
\item We build a Style-driven Multi-scale Adaptive Refinement Transformer to refine the intermediate style features of the generator so that high-quality inverted images can be attained, which can also surprisingly adapt to editing tasks.
\item We explore the disentanglement of $\mathcal{W^+}$ and editability of $\mathcal{F}$, further conduct qualitative and quantitative experiments to prove the superiority of StylePrompter that achieves a balance between quality and editability.

\end{itemize}
\begin{figure*}
  \includegraphics[width=\textwidth]{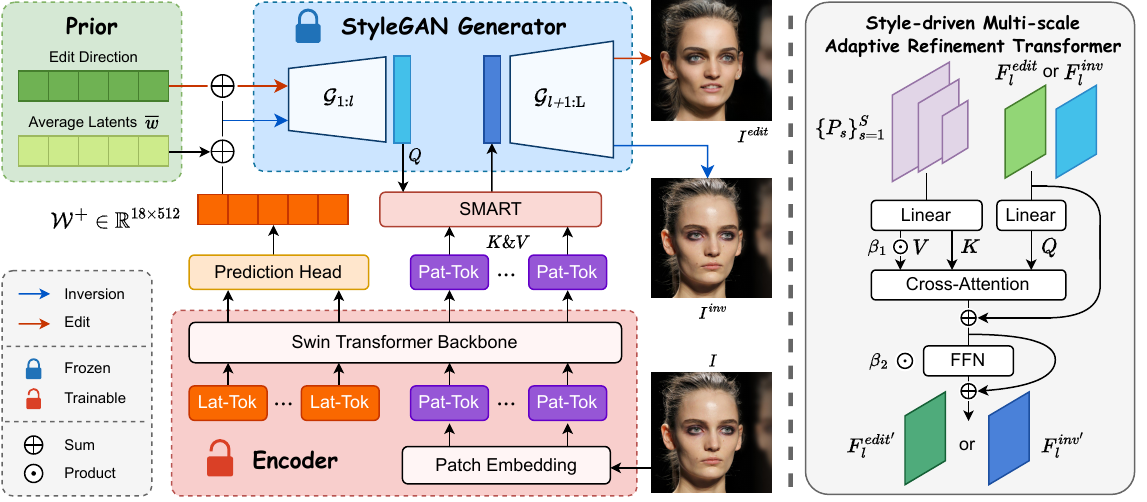}
  \caption{Overall StylePrompter architecture. We transfer Swin Transformer backbone for GAN inversion where latent codes are embedded as latent tokens (Lat-Tok), interacting with the patch tokens (Pat-Tok) of a flattened image to extract information at both image level and latent level. The SMART block is designed to further refine the intermediate style features in the StyleGAN generator. Style features are treated as queries, fetching lost identity information from the encoder's multi-scale features. SMART can also adapt to edited style feature maps, "smart" just as its name.}
  \vspace{-0pt}
  \Description{Enjoying the baseball game from the third-base
  seats. Ichiro Suzuki preparing to bat.}
  \label{method}
\end{figure*}

\section{Related Works}
\subsection{GAN Inversion}
Given a pre-trained GAN model, GAN inversion aims to find the most appropriate latent codes to represent the input images. There are typically three categories: optimization-based, learning-based, and hybrid. Optimization-based approaches \cite{abdal2019image2stylegan, abdal2020image2stylegan++, zhu2020improved} start from randomly initialized latent codes updated by minimizing the reconstruction error per image through gradient descending. Learning-based methods train encoders to deal with a collection of images. Hybrid methods \cite{roich2022pivotal, mao2022cycle} combine both. Generally, inversion via optimizing can achieve high-fidelity images but suffer from long inference time and unstable edits. In contrast, encoders are capable of inferring in a short time, efficiently embedding images into latent codes, which will be of more practical significance. 

Previous learning-based works \cite{richardson2021encoding, tov2021designing, alaluf2021restyle, wei2022e2style, hu2022style} focus on predicting latent codes in $\mathcal{W^+}$, which has been proven to be the best space to trade-off for image quality and editing flexibility \cite{kang2021gan, roich2022pivotal}. To better alleviate distortion, recent works \cite{yao2022feature, wang2022high, alaluf2022hyperstyle, dinh2022hyperinverter, bai2022high} apply a two-stage strategy, utilizing the above $\mathcal{W^+}$-based encoder to get coarse inverted images, and further add information in deeper latent space. 
As most methods carry out the work based on CNNs, our method instead resorts to Transformer-based model, exploring the effectiveness of the attention mechanism for GAN inversion.

\subsection{Latent Space Manipulation}
Image editing has been a long-standing open research problem. Unlike per-pixel image editing algorithms, StyleGAN provides a semantically rich latent space that can realize diverse image manipulation at latent level. Numerous works have explored this disentanglement latent space to identify semantic directions in a supervised \cite{abdal2021styleflow, goetschalckx2019ganalyze, shen2020interpreting}, unsupervised \cite{harkonen2020ganspace, shen2021closed, voynov2020unsupervised} or self-supervised \cite{plumerault2020controlling, jahanian2019steerability} manner. Recently, CLIP \cite{radford2021learning} sparks research on text-based image manipulation, of which the text and image encoders are investigated to perform various unsupervised semantic edits \cite{abdal2022clip2stylegan, patashnik2021styleclip}.

The above works facilitate GAN inversion in latent space manipulation. Such editing directions have been found to affect different layers in the StyleGAN generator. For example, the pose is mainly controlled by shallow layers, while a change in hairstyle is closely related to deeper convolution layers. Existing $\mathcal{F}$-involved methods fail to balance inversion quality and edits due to their strict manner of refinement in deeper latent space, while our proposed SMART will not deteriorate the editability of $\mathcal{F}$ space.

\subsection{Transformer in GAN Inversion}
Transformer \cite{vaswani2017attention} has achieved unprecedented performance, and started to make an impact in both natural language processing (NLP) and computer vision. In NLP it has become a state-of-the-art method that fine-tunes pre-trained Transformer models on token-level tasks. This idea has been successfully adopted to tackle the computer vision tasks such as object detection upon the Vision Transformer (ViT) in a pure sequence-to-sequence learning manner \cite{fang2021you, dosovitskiy2020image}. We then raise the following question: \textbf{Can a Transformer-based model transfer to GAN inversion at token level?} 

Two works \cite{hu2022style, liu2022delving} have explored the Transformer module for GAN inversion, however, both build on a CNN-based backbone. To our knowledge, we are the first to adopt such a vision Transformer model as backbone.
Standard ViT models use tokens with fixed lengths, which leads to the loss of image-level information necessary for reconstruction. Some hierarchical vision Transformer models, such as Swin Transformer \cite{liu2021swin, liu2022swin}, following the traditional technique of CNNs, introduce a multi-scale architecture, which has different sizes of feature maps among blocks. Motivated by previous works, we transfer a hierarchical vision Transformer model for GAN inversion, encouraging the encoder to produce fine-grained latent codes at token level as well as feature maps at image level.

\section{Method}
An overview pipeline of StylePrompter architecture is illustrated in Figure \ref{method}. We adopt Swin Transformer to tackle this quality-editability trade-off. We first predict $\mathcal{W^+}$ latent codes at token level and then retrieve lost identity information in $\mathcal{F}$ space.
\subsection{Transformer-based Encoder}
Inspired by \cite{dosovitskiy2020image, fang2021you}, we proposed a novel encoder framework for GAN inversion, fine-tuning a pre-trained Transformer-based backbone to predict $\mathcal{W^+}$ latent codes at token level and extract image-level features.
In practice, we choose Swin Transformer since it performs a pyramidal feature extraction like CNNs, which can provide multi-scale image features adequate for refinement in SMART (Section 3.2). To distinguish the feature maps from the encoder and decoder, we use $P$ to represent \textit{Pyramidal} features extracted from the encoder and $F$ as style \textit{Feature} maps of the generator. 

Normally, images are embedded as patch tokens in standard vision Transformer models. To extend to GAN inversion, we append $T$ randomly initialized latent tokens which are learnable during training time. However, it is impossible to naively concatenate these additional latent tokens with patch tokens as the input of the backbone like \cite{dosovitskiy2020image, fang2021you}, because the carefully designed window partition operation in Swin will divide several patches into a window before self-attention, which should accomplish at image level.

\begin{figure}[h]
  \centering
  \includegraphics[width=\linewidth]{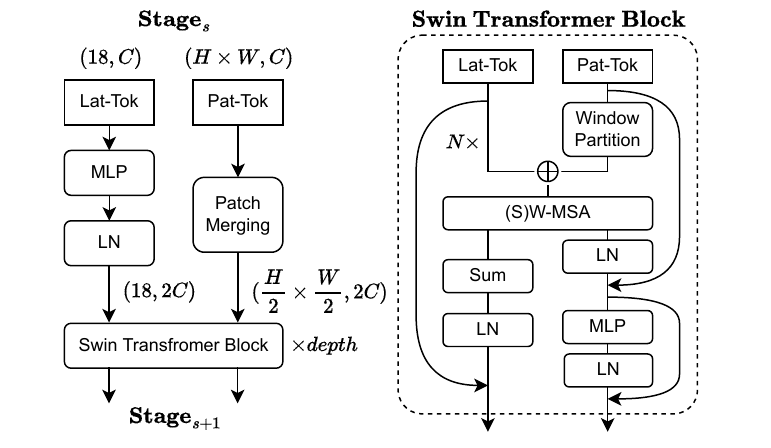}
  \caption{Latent tokens involved Swin Transformer. Latent tokens (Lat-Tok) replicate several times and concatenate with partitioned Patch Tokens (Pat-Tok) as the input of (S)W-MBA. Latent tokens only interact with patch tokens in attention.}
  \vspace{-10pt}
  \Description{Revised Lat-Tok participated Swin backbone.}
  \label{latent-tokens}
\end{figure}

Here we propose a novel token-involved fine-tuning approach for hierarchical vision Transformers. 
To encourage latent tokens to participate in every window, we first replicate latent tokens $N$ times, where $N$ is the number of windows. Each replication of latent tokens will be concatenated with a window of partitioned patch tokens as the input of (Shifted) Window-based Multi-head Self-attention block. As illustrated in Figure \ref{latent-tokens}, after self-attention, patch tokens will follow the normal routine that pass to a 2-layer MLP followed by LayerNorm (LN) layer while the repeated latent tokens will be back to the original shape through summation before normalization, and no MLP applied. Between stages, the number of patch tokens reduces (\begin{math}2\times\end{math} downsampling of resolution) through patch merging, while for latent tokens we only apply a simple MLP layer to match the dimension with patch tokens. The output latent tokens will finally be projected to $\mathcal{W^+}$ space by a 3-layer MLP with Tanh activation in between as the prediction head. We follow pSp \cite{richardson2021encoding} to learn a residual of average latent codes in StyleGAN prior, denoted as \begin{math}\overline{w}\end{math}. More details can be found in Appendix. To be simple, we formulate the predicted latent codes of inverted images as:
\begin{equation}
\label{inversion_equation}
w^{inv}=MLP(E(X)) + \overline{w}
\end{equation}
where $E$ is the Swin backbone, and MLP is the prediction head. 

Notice that only in self-attention will latent tokens interact with patch tokens, where increased complexity can be negligible. As standard multi-head self-attention (MSA) is quadratic to the patch number, window-based multi-head self-attention (W-MSA) is linear when the window size $M$ is fixed. Our revised one, denoted as W-MSA*, can approximately be the same as W-MSA. To make the cost clear, we formulate the computational complexity of the above on an image of \begin{math}h\times w\end{math} patches as follows:
\begin{equation}
\Omega(MSA) = 4hwC^2+2h^2w^2C
\end{equation}
\begin{equation}
\Omega(W\mbox{-} MSA) = 4hwC^2+2M^2hwC
\end{equation}
\begin{equation}
\Omega(W\mbox{-} MSA^*) = 4(hw+T)C^2+2M^2(hw+T)C
\end{equation}
where $T$ is the number of appended latent tokens.
\subsection{Style-driven Multi-scale Adaptive Refinement Transformer}
Latent codes in $\mathcal{W^+}$ can only represent a coarse inversion that is not faithful enough for real-world tasks. To improve the quality of the inverted images, we carefully design a so-called Style-driven Multi-scale Adaptive Refinement Transformer (SMART) block to modify the intermediate style feature maps in the generator through the cross-attention mechanism. 

Let $F_l=G(w_{1:l})$ denote the output feature maps at the $l$-th convolution layer of the StyleGAN generator, which is controlled by latent codes $w_{1:l}=\{w_1,...,w_l\}$. The multi-scale image feature maps extracted by the encoder are represented as $P=\{P_s\}^S_{s=1}$, where $S$ is the stage number. Suppose that some specific identity information can be lost in $F_l$ caused by limited expressiveness of $\mathcal{W^+}$, we alight on the idea of treating style features as query elements by linear projection to retrieve the lost information from key and value elements, which are linear projections of $P$. 
Moreover, as $Q$, $K$, $V$ are projected from which are of pixels, we naturally utilize local attention to constrain each query to look at the key and value elements in the same spatial location, therefore retrieving accurate spatial information and be more efficient in computation. More details are provided in Appendix. We do emphasize that it is not a common scaled dot-product attention, since we skip the Softmax and scaled operation when computing the dot product of query and key. We incline to use the dot production of $Q$ and $K$ to quantify the missing identity of style features instead of finding the correlation between the two components. We formulate the modified style features as follows:
\begin{equation}
\label{cross-attention_inv}
\hat F'_l=Attention(Q,K,V)+F_l
\end{equation}
\begin{equation}
\label{FFN_inv}
F_l'=MLP(\hat F'_l)+\hat F'_l
\end{equation}
where \begin{math}MLP(\hat F'_l)=max(0,\hat F'_l W^1+b^1)W^2+b^2\end{math}. Notice that we did not apply any Norm layer since we found in practice that the normalization will hurt the capacity of controlling manipulation. 
After all, $F'_l$ will replace the original feature map $F_l$ and feed into the generator again to achieve a high-quality inverted image. 

Simply, the refined feature maps are denoted as:
\begin{equation}
    F'_l=SMART(F_{l},\{P_s\}^S_{s=1})
\end{equation}

It is not new to modulate codes in $\mathcal{F}$ space. However, instead of directly replacing the original style features \cite{yao2022feature} or refining via affine transformation \cite{wang2022high}, our proposed SMART take full advantage of prior style information in a \textit{smart} manner which can not only fix the destroyed style feature maps for better inversion quality but also adapt to the edited style features in editing task. 

\subsection{Latent Manipulation}
The disentangled latent space of StyleGAN offers editability for attributes-based latent manipulation. Given an input image, we first obtain its corresponding latent codes $w^{inv}$ by Equation (\ref{inversion_equation}). The editing direction $\Delta w$ concerning specific attributes can be obtained from off-the-shelf methods as prior knowledge. Thus we can manipulate the latent codes as $w^{edit}=w^{inv}+ \alpha \cdot \Delta w$, where $\alpha$ is the magnitude of manipulation.

Let $F^{edit}_l=G(w^{edit}_{1:l})$ denotes the changed $l$-th style feature maps in the generator. We modify the Equation (\ref{cross-attention_inv})(\ref{FFN_inv}) as follows:
\begin{equation}
\hat F^{edit'}_l=Attention(Q,K,\beta_1 \cdot V)+F^{edit}_l
\end{equation}
\begin{equation}
F^{edit'}_l=\beta_2 \cdot MLP(\hat F^{edit'}_l)+\hat F^{edit'}_l
\end{equation}
where $Q$ is a projection of $F^{edit}_l$, $K$ and $V$ are the same as inversion. Furthermore, except for the basic magnitude $\alpha$ controlling the strength of editing, we encourage an auxiliary weight \begin{math}\beta=(\beta_1, \beta_2)\end{math} to involve at two residual connections. For some edits allowed to be inconsistent with the input image, e.g., age changing, we use a smaller $\beta_1$ to reduce the rate of retrieved identity information from value elements in SMART, providing more flexibility for edits. 

Simply, the modified style features can be expressed as:
\begin{equation}
F_l^{edit'}=SMART(F_l^{edit},\{P_s\}^S_{s=1}, \beta)
\end{equation}
where $\beta=(\beta_1, \beta_2)$ is fixed to $(1, 1)$ during training.

\subsection{Training Objectives}
For training, we use a two-stage strategy that first trains the backbone and prediction head to get $\mathcal{W^+}$ latent codes, then fine-tune SMART until converges. To ensure a high-quality reconstruction, we follow the protocol of previous works to apply the pixel-wise $L_2$ loss, LPIPS perceptual loss \cite{zhang2018unreasonable}, and identity similarity loss which calculates the cosine similarity between two image embeddings. For the face domain, we employ a pre-trained face recognition network ArcFace \cite{deng2019arcface} to preserve facial identity. 
Let $I, I^{inv}$ denote the input image and the inverted one, and the image loss is defined as:
\begin{equation}
\label{image_loss}
L_{image}=\lambda_{1}L_{2}(I, I^{inv})+\lambda_{2}L_{lpips}(I, I^{inv})+\lambda_{3}L_{id}(I, I^{inv})
\end{equation}
where $\lambda_1,\lambda_2,\lambda_3$ are the hyper-parameters.

When training the prediction head, we follow pSp \cite{richardson2021encoding} to align the predicted latent codes with the average latent vector $\overline{w}$, enforcing proximity to the center of $\mathcal{W}$ space through $L_2$ regularization:
\begin{equation}
\label{align_loss}
L_{align}=||w^{inv}-\overline{w}||_2
\end{equation}

The total loss for the baseline can be represented as follows:
\begin{equation}
L_{base}=L_{image}+\lambda_{4}L_{align}
\end{equation}
where $\lambda_4$ is the hyper-parameter.

When training SMART, we only employ the common image loss as Equation (\ref{image_loss}).
\begin{figure*}
  \includegraphics[width=\textwidth]{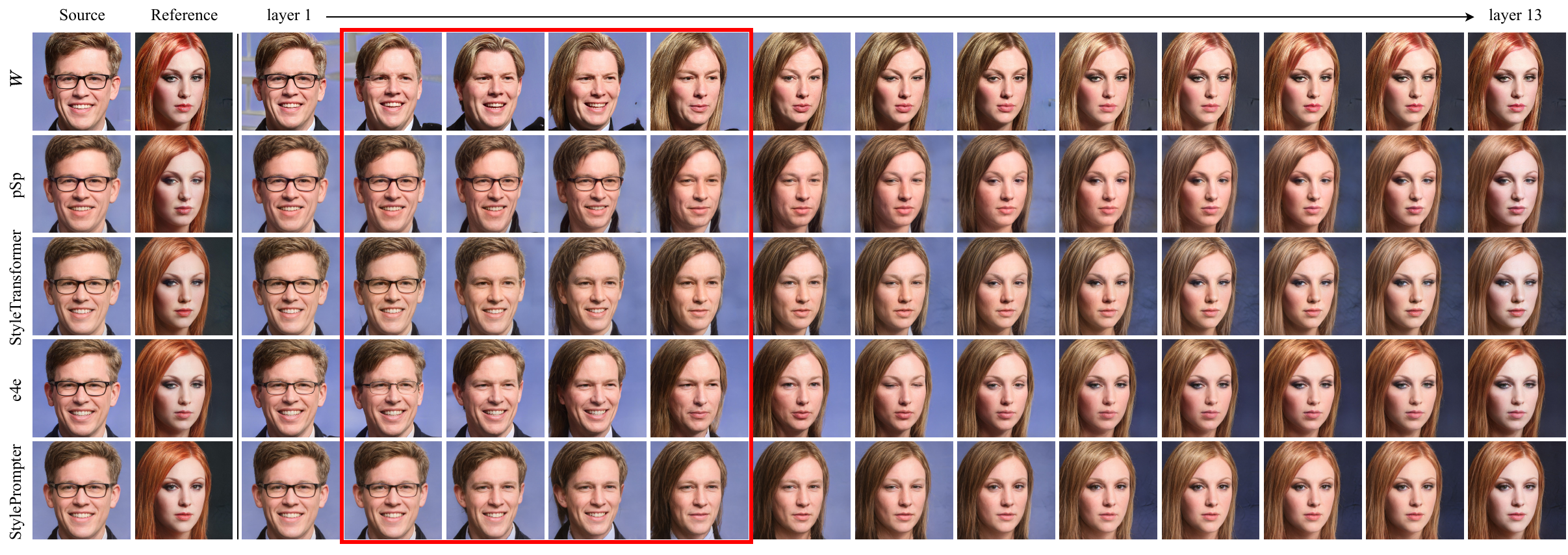}
  \caption{We utilize style mixing to investigate the disentanglement visually. By progressively replacing the latent codes of the source image with that of the reference image, the intermediate image can change gradually. Compared with the original $\mathcal{W}$ and other $\mathcal{W^+}$-based inversion methods, StylePrompter gives the cleanest and sharpest variations of eyeglasses, pose and smiling illustrated by the red boxes, indicating more disentanglement.}
  \Description{Disentanglement of $\mathcal{W^+}$.}
  \label{disentanglement_comparison}
\end{figure*}
\section{Experiments}
\subsection{Settings}
\textbf{\textit{Configurations and Datasets}}.
We apply a two-stage strategy that first fine-tunes a pre-trained Swin Transformer, SWINv2-T \cite{liu2022swin} removing the classification head to be specific, then fix it to train SMART until converges. The generator is frozen for both stages. Ranger optimizer is used to update trainable parameters, which combines Rectified Adam \cite{liu2019variance} with the Lookahead technique \cite{zhang2019lookahead}. The learning rate and batch size are $1\times 10^{-3}$ and 4 for both stages. For human face domain reconstruction, we employ the FFHQ \cite{karras2019style} dataset with 70k high-quality for training, and evaluate on the first 1k images of CelebA-HQ \cite{karras2017progressive}. 
\\
\textbf{\textit{Baselines}}.
We focus on learning-based methods for comparison. Our base model (without SMART, marked with an asterisk) will be compared with classic methods pSp \cite{richardson2021encoding}, e4e \cite{tov2021designing}, and the state-of-the-art method StyleTransformer \cite{hu2022style} which are $\mathcal{W^+}$-based encoders. Our full model will be compared with HFGI \cite{wang2022high}, and FeatureStyleEncoder \cite{yao2022feature} which also participate in $\mathcal{F}$ space.
\\
\textbf{\textit{Metrics}}.
We estimate the performance of different methods based on the quality-editability trade-off, which should be evaluated from the aspects of fidelity and realism. To be specific, we use the full reference metrics $L_2$ and LPIPS \cite{zhang2018unreasonable} to evaluate fidelity, calculating pixel-wise and perceptual similarity between image pairs. For realism, we employ a no-reference image quality assessment metric MANIQA \cite{yang2022maniqa} which is artifact sensitive, instead of the widely-used FID \cite{heusel2017gans}. To quantify the identity preservation of edited images, we employ another face recognition network Curricularface \cite{huang2020curricularface}, instead of ArcFace \cite{deng2019arcface} used in training.
\subsection{Interpretation and Explanation}
\textbf{\textit{Which $\mathcal{W^+}$ is more disentangled?}}
Different from \cite{wu2021stylespace} that explores the disentanglement between different latent spaces, we focus on comparing inversion methods in the same latent space, $\mathcal{W^+}$. Review the hypothesis in e4e \cite{richardson2021encoding} that latent codes closer to $\mathcal{W}$ space correspond to images that are less faithful but more realistic and better editability, which we doubt since $\mathcal{W}$ has been proven to be entangled, causing aliasing. In this part, we attempt to give explanations for entanglement and offer some novel insights.
\begin{table}
  \caption{The dispersion degree and distance to $w$ for different $W^+$-based methods. LPIPS and MANIQA scores as a reference.}
  \label{disentanglent}
  \resizebox{\linewidth}{!}{
  \begin{tabular}{ccccc}
    \toprule
    Methods & Dispersion & Distance & LPIPS$\downarrow$ & MANIQA$\uparrow$ \\
    \midrule
    pSp & 9.570 & 1025.5 & 0.131 & 0.0182 \\
    e4e & 0.079 & 156.9 & 0.151 & 0.0212 \\
    StyleTransformer & 0.886 & 438.5 & 0.127& 0.0182 \\
    StylePrompter & 0.066 & 181.4 & 0.139 & 0.0194 \\
    \bottomrule
  \end{tabular}
  }
  \vspace{-10pt}
\end{table}

As a beginning, we estimate the property of $\mathcal{W^+}$ numerically from two aspects: the dispersion degree and distance to $\mathcal{W}$. More specifically, we randomly generated 1,000 images using a pre-trained StyleGAN generator which $w\in \mathcal{W}$ is known, then obtain the corresponding codes $\hat w=\{w_l\}_{l=1}^{L}$ by pSp, e4e, StyleTransformer and StylePrompter (without SMART), calculating the average standard deviation (std) of $\hat w$ and the average Manhattan distance between $w$ and $\hat w$. On the one hand, a lower dispersion degree indicates that latent codes are concentrated at one point, i.e., closer to $\mathcal{W}$. On the other hand, the distance demonstrates whether the correct value of attributes is predicted, i.e., equal to $w$. As listed in Table \ref{disentanglent}, e4e and our method both are closer to the original $\mathcal{W}$ space than StyleTransformer, while pSp completely deviates. However, it is strange that latent codes predicted by StylePrompter bound together but are not close to $w$, as we get a lower score in dispersion degree but higher in distance than e4e, encouraging us to further study the correlation of latent codes between layers.
\begin{figure}
  \includegraphics[width=\linewidth]{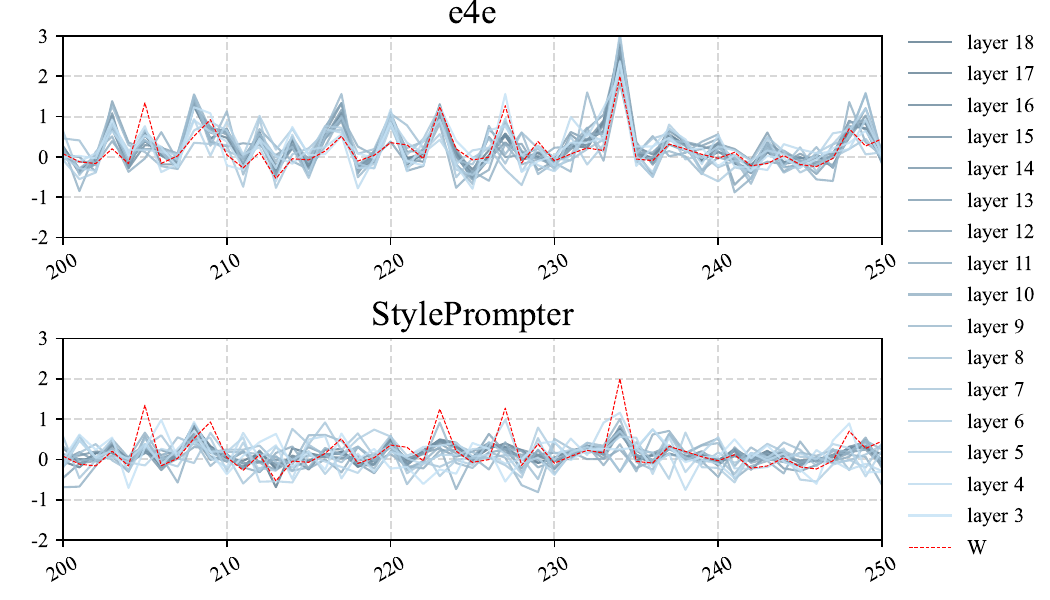}
  \caption{The relationship between $w$ and $\hat w$ per layer. We visualize the value of channels 200-250 for one generated sample. It reveals that channels of all layers in e4e are aligned with $w$ but only a few in StylePrompter are closely related.}
  \Description{Enjoying the baseball game from the third-base
  seats. Ichiro Suzuki preparing to bat.}
  \label{respond}
  \vspace{-16pt}
\end{figure}

It has been verified that the impact on attributes is different among layers. Thus we raise the following presumption: \textbf{Different layers respond to specific attributes in varying degrees}. To find more evidence, we plot the inverted latent codes $\hat w$ from e4e and StylePrompter to compare with $w$. pSp and StyleTransformer are skipped since they are not aligned with $\mathcal{W}$. Figure \ref{respond} indicates that although the two methods are both related to $w$, layers in e4e mostly are close to $w$ while only a few are for StylePrompter. 

Let us assume StylePrompter lies in a more disentangled $\mathcal{W^+}$ where attributes are correctly learned for certain layers, and we further explore what it will affect through three kinds of style mixing: \textit{progressively replacing}, \textit{one-layer exchanging} and \textit{interpolation}. We claim that the variation of attributes concentrating on fewer layers indicates more disentanglement. Results of progressively replacing are shown in Figure \ref{disentanglement_comparison}. The pose changed at more than one layer in the entangled $\mathcal{W}$ space, and nonsense textures appear in the background. In a similar case, intermediate results produced by e4e conform neither to the source image nor the reference image (gazing direction), demonstrating attributes entanglement. In contrast, StylePrompter gives the sharpest and cleanest change than other methods, revealing that it is capable of disentangling attributes among layers and allocating more values where responses are more active. The results of one-layer exchanging and interpolation provided in Appendix are also in line with this finding.

For the reason why StylePrompter can find a more disentangled $\mathcal{W^+}$, we suppose that owe to the participation of latent tokens in feature extraction that is aware of how attributes are recognized, while other methods only utilize the output features for prediction. Self-attention is also a reason that latent tokens can communicate with each other. Therefore, we can consolidate and complement the theory from e4e as follows: (\romannumeral1) Widely dispersed $w$ codes produce unreal textures in the inverted images; (\romannumeral2) Be close to $\mathcal{W}$ space in every layer resulting in entanglement may not be helpful for editing. 
It is our belief that a more disentangled $\mathcal{W^+}$ can further benefit the interpretation of the latent spaces in StyleGAN.
\begin{figure}
  \centering
  \includegraphics[width=\linewidth]{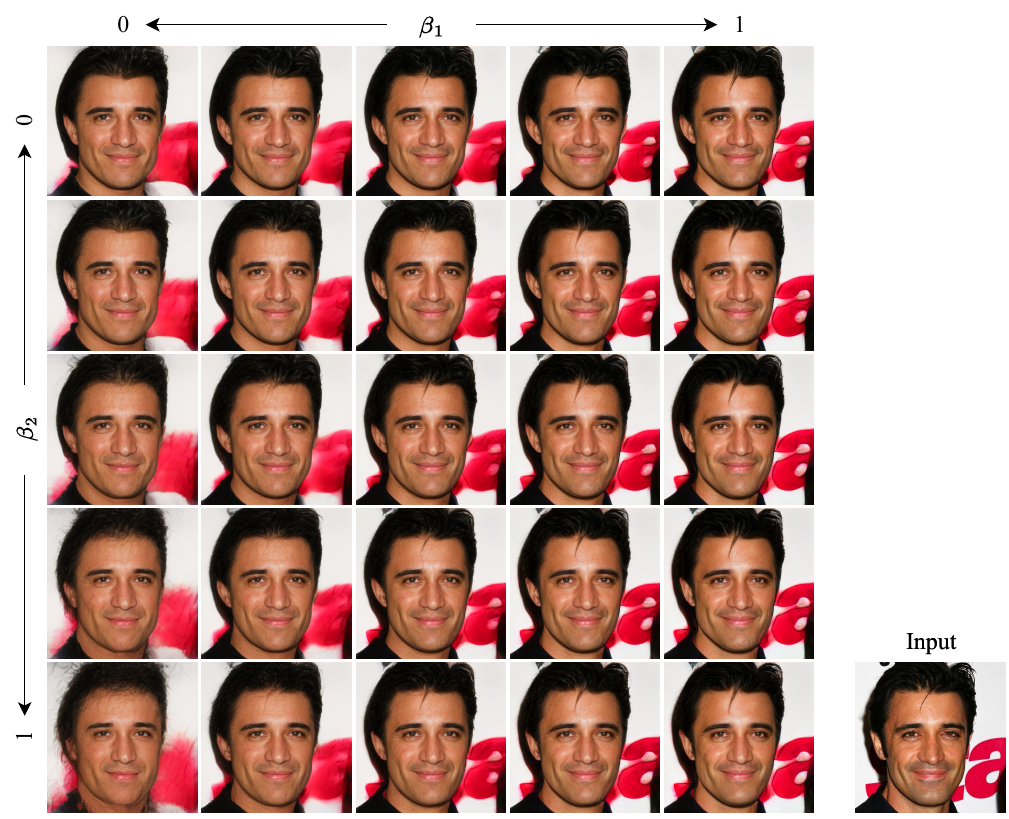}
  \caption{Inversion results via StylePrompter with SMART in different $\beta$. It demonstrates that the cross-attention learns to "add" identity information while FFN learns to "subtract" invalid values or unimportant information.}
  \Description{Details of Multi-scale Attention.}
  \label{beta}
  \vspace{-12pt}
\end{figure}
\begin{figure}
  \centering
  \includegraphics[width=\linewidth]{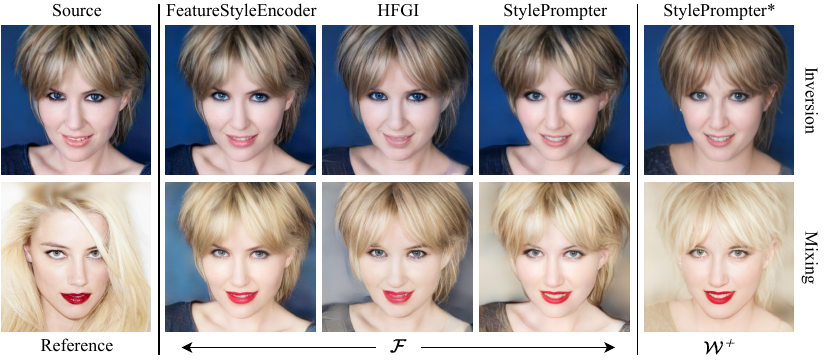}
  \caption{Style mixing via $\mathcal{F}$-involved inversion methods. Refinements in FeatureStyleEncoder and HFGI are too strict to keep the editability of $\mathcal{F}$, while StylePrompter in a soft manner will not weaken this ability, and can produce a manipulated image much closer to what is completed in $\mathcal{W^+}$.}
  \Description{Editability in $\mathcal{F}$.}
  \label{disentanglement_comparison_f}
  \vspace{-12pt}
\end{figure}
\\
\textbf{\textit{Why SMART?}}
Another important component of StylePrompter is SMART. In this part, we aim at understanding what SMART has learned. To be specific, we manipulate the additional weights $\beta$, which are designed to control the residual values. The result in Figure \ref{beta} demonstrates that increased $\beta_1$ with respect to cross-attention can retrieve more identity information while increasing $\beta_2$ with respect to FFN will ignore the unimportant patterns like background. It suggests that the cross-attention in SMART learns to \textit{add} information, and FFN learns to \textit{subtract} invalid value caused by the previous residual connection. With the controllable SMART, we can provide more flexibility for edits.

To compare with other $\mathcal{F}$-involved inversion methods, FeatureStyleEncoder does \textit{hard} refinement that directly replaces the original style features, HFGI that modifies features via affine transformation is \textit{relatively soft}, while SMART using residual connections with controllable weights is indeed \textit{soft}. In Figure \ref{disentanglement_comparison_f} we utilize style mixing again to visualize the effects caused by different refinement manners. We replace the $\mathcal{W^+}$ latent codes of the source image with that of the reference image after layer 7, where all the above methods have finished refinement. The fine details should have been controlled by deeper layers, but HFGI and FeatureStyleEncoder are incapable of removing the styles of the source image, especially for the wrinkle, eyes, and dark blue background, while StylePrompter maintains the editability of $\mathcal{F}$ space, producing manipulated image more similar to the reference one.
\begin{figure*}
  \includegraphics[width=\textwidth]{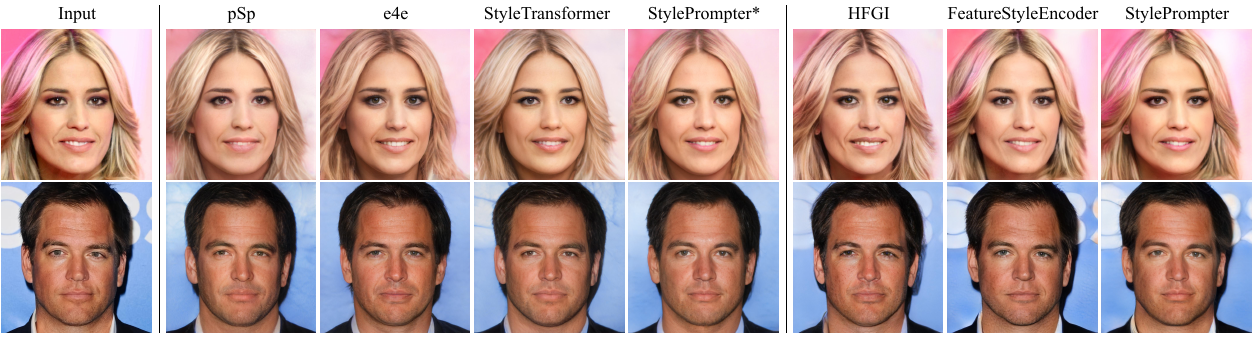}
  \caption{Qualitative comparison for encoder-based GAN inversion methods. $\mathcal{W^+}$-based methods (columns 2-5) suffer from the reconstruction of details, while $\mathcal{F}$-involved methods (columns 6-8) succeed in preserving more identity information. StylePrompter (columns 5 and 8) can be comparable to the state-of-the-art methods.}
  \Description{Enjoying the baseball game from the third-base
  seats. Ichiro Suzuki preparing to bat.}
  \label{inversion_results}
\end{figure*}
\subsection{Inversion Results}
We then conduct an overall comparison for StylePrompter with both $\mathcal{W^+}$-based methods and $\mathcal{F}$-involved methods.
\\
\textbf{\textit{Qualitative Evaluation}}. As shown in Figure \ref{inversion_results}, $\mathcal{W^+}$-based methods suffer from limited expressiveness and fail in preserving identity. Exploiting more expressive latent space, $\mathcal{F}$-involved methods can visually outperform $\mathcal{W^+}$-based methods with finer structural details. Applying hard refinement, FeatureStyleEncoder achieves the best inversion quality, while HFGI and ours fail to reconstruct some out-of-domain details. Notice that facial artifacts can appear in the inversion results of HFGI, e.g., teeth, which is crucial for realism, as it modifies style feature maps based on the difference between the coarse reconstruction image and the input one, in the case of excessive mismatch will artifacts appear. In contrast, StylePrompter learns a residual on the prior style feature maps, and will not deviate from the original distribution. Moreover, the effectiveness of SMART can be observed in completed image details like the background, compared with StylePrompter*.
\\
\textbf{\textit{Quantitative Evaluation}}. As listed in Table \ref{quantitative_evaluation}, metrics on $\mathcal{W^+}$-based encoders have little difference. e4e gets the highest MANIQA scores as it predicts latent codes strictly approaching the $\mathcal{W}$ space. StylePrompter without SMART is able to reconcile fidelity with realism, getting moderate scores in all metrics. While appending SMART, our full model can outperform $\mathcal{W^+}$-based methods in the metrics of fidelity with less than 4M parameters added, demonstrating the effectiveness and efficiency of SMART. Among $\mathcal{F}$-involved methods, StylePrompter overtakes HFGI but is slightly inferior to FeatureStyleEncoder. However, we do emphasize that FeatureStyleEncoder has a notable limitation in editing. We next compare the editability in the following subsection, showing the adaptability of StylePrompter for editing tasks. 
\begin{table}
  \caption{Quantitative evaluation of encoder-based methods measured on CelebA-HQ, the first 4 rows are $W^+$-based and the last 3 rows are $\mathcal{F}$-involved methods. The best and runner-up are marked in bold and underline, respectively.}
  \label{quantitative_evaluation}
  \resizebox{\linewidth}{!}{
  \begin{tabular}{ccccc}
    \toprule
    Methods & $L_2$$\downarrow$ &LPIPS$\downarrow$& MANIQA$\uparrow$& Params(M)$\downarrow$ \\
    \midrule
    pSp                 & \underline{0.040} & \underline{0.153} & 0.0170 & 297.50\\
    e4e                 & 0.052 & 0.189 & \textbf{0.0204} & 297.50\\
    StyleTransformer                  & \textbf{0.039} & \textbf{0.150} &  0.0170 & \underline{70.99}\\
    StylePrompter*                 & 0.041 & 0.164 & \underline{0.0188} & \textbf{60.70} \\
    \hline
    HFGI                & 0.027 & 0.111 & 0.0166 & 303.75\\
    FeatureStyleEncoder                 & \textbf{0.019} & \textbf{0.062} & \textbf{0.0203} & \underline{113.92}\\
    StylePrompter                  & \underline{0.022} & \underline{0.089} & \underline{0.0175} & \textbf{64.44} \\
    \bottomrule
  \end{tabular}}
\end{table}
\begin{figure*}
  \includegraphics[width=\textwidth]{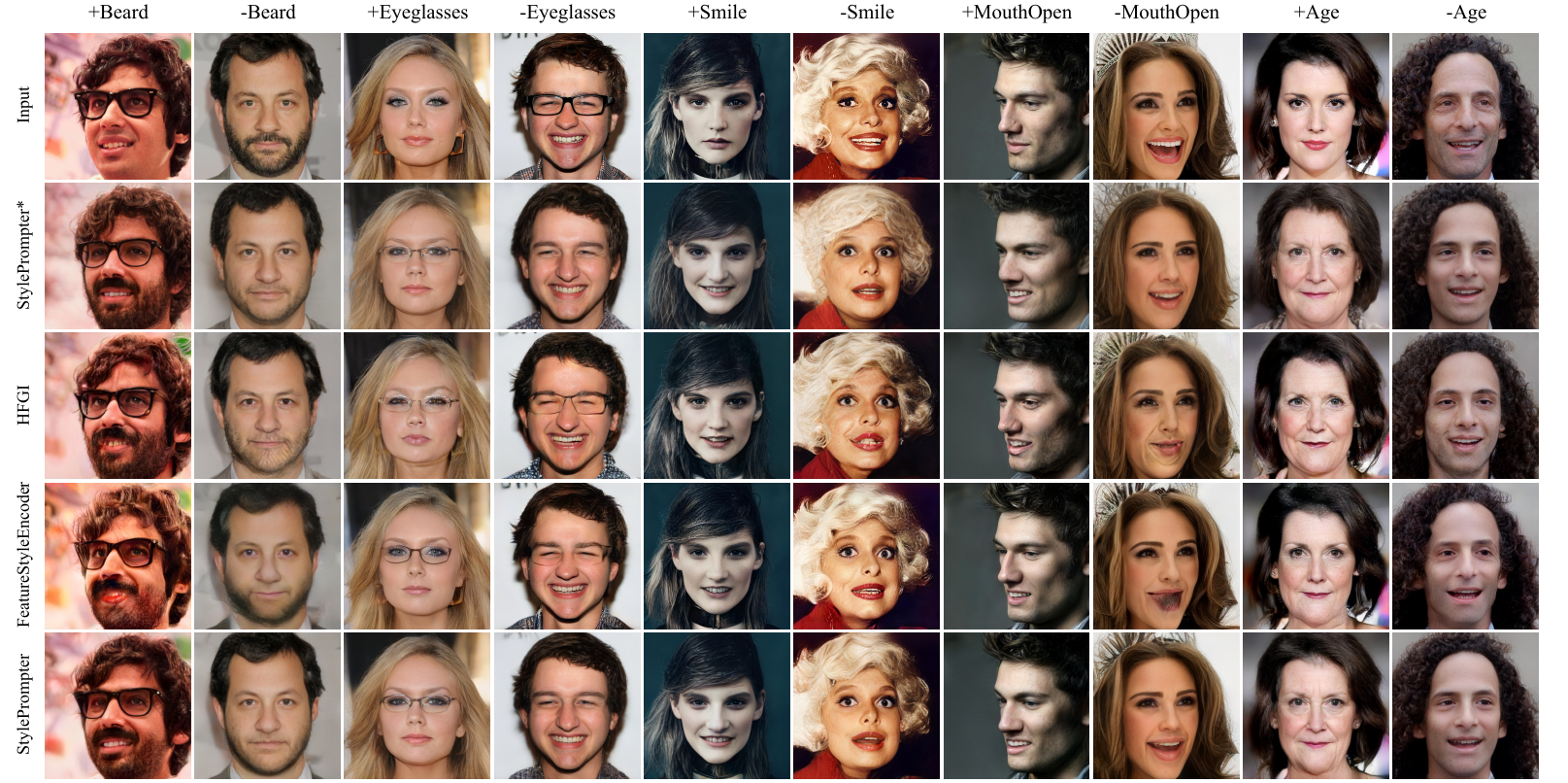}
  \caption{Editing results via different $\mathcal{F}$-involved methods and StylePrompter without SMART as a reference. In most cases, HFGI and FeatureStyleEncoder fail to produce faithful and meaningful results, especially in the edits of "minus", such as removing eyeglasses and closing the mouth. The reason lies in their manners of modification that are not flexible enough to maintain the editability of $\mathcal{F}$ space. StylePrompter however providing variable refinement can adapt to most edits.}
  \Description{Enjoying the baseball game from the third-base
  seats. Ichiro Suzuki preparing to bat.}
  \label{editing_comparison}
\end{figure*}
\subsection{Editing Results}
\begin{figure}
  \includegraphics[width=\linewidth]{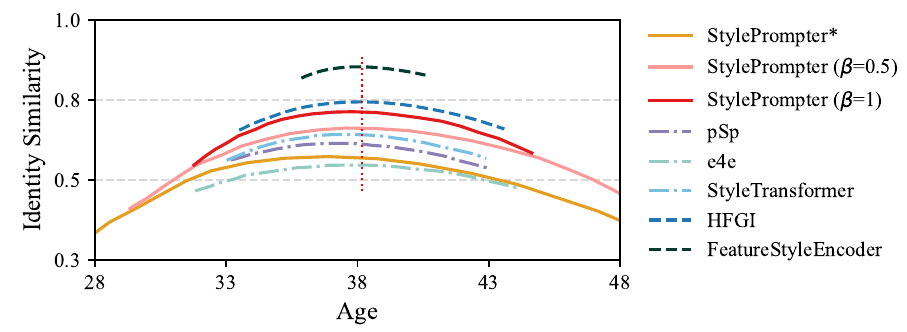}
  \caption{Age manipulation in the same range of magnitude $\alpha$ for different inversion methods. StylePrompter with disentangled $\mathcal{W^+}$ is the most changed, and further with SMART can balance edits and identity consistency.}
  \Description{Details of Multi-scale Attention.}
  \label{age_estimation}
\end{figure}

\textbf{\textit{Qualitative Evaluation}}. The principle of GAN inversion is to edit images toward target attributes and maintain identity consistency. Here we focus on comparing the editability between $\mathcal{F}$-involved methods. We also provide the editing results of our base model for reference and leave the comparison with other $\mathcal{W^+}$-based methods in Appendix. All the editing directions are obtained from \cite{abdal2022clip2stylegan, patashnik2021styleclip}. Notice that we control the edits visually with the same strength of target attributes, which means $\alpha$ may be different among methods. Figure \ref{editing_comparison} shows that $\mathcal{W^+}$ has better editability but fails in persevering identity. FeatureStyleEncoder and HFGI cannot generate desirable edited images, especially when removing attributes, caused by inappropriate refinement manners. In contrast, StylePrompter is "smart" enough to fit any edit case. Furthermore, we can loosen the identity with the help of smaller $\beta$ to achieve more flexible edits.
\\
\textbf{\textit{Quantitative Evaluation}}.
To quantitatively evaluate the editability, we involve an off-the-shelf model, coral \cite{cao2020rank}, to estimate the changes of age and the preservation of identity by Curricularface \cite{huang2020curricularface}. The comparison results are plotted in Figure \ref{age_estimation}. 
Intuitively, the curve in a wider range shows better editability, yet a higher ID score indicates the robustness of manipulation. Although achieved remarkable inversion results, FeatureStyleEncoder shows poor editability. StylePrompter without SMART exhibits the largest variation among $\mathcal{W^+}$-based methods, proving its disentanglement. Appending SMART, StylePrompter can preserve more identity and maintain editability at the same time. Notice that the excessive manipulation would influence the measurement, resulting in dramatically descended ends of the curve.

Recall that the main goal of GAN inversion is to strike a balance between maintaining identity consistency and editing flexibly. The above comparison highlights the brilliant property of StylePrompter, which conforms to this goal. 
\begin{figure}
  \includegraphics[width=\linewidth]{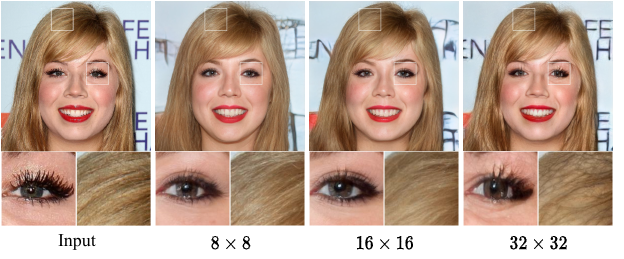}
  \caption{SMART in higher resolution ($32\times 32$) with finer details but can produce unrealistic artifacts, while lower resolution ($8\times 8$) is not faithful enough.}
  \Description{Details of Multi-scale Attention.}
  \label{ablation}
\end{figure}

\subsection{Ablation Study}
The choice of the intermediate $l$-th convolution layer is crucial for SMART. We first decide which resolution of feature maps can better balance fidelity and realism. More specifically, we train three different models that locate SMART at the layer with respect to resolutions of $8\times 8$, $16\times 16$, and $32\times 32$. We visualize the images inverted by each configuration in Figure \ref{ablation}. Although the refinement in the higher layer can invert images more faithfully, even able to reconstruct out-of-domain details (letters in the background), it suffers from heavy artifacts. While SMART at resolution $8\times 8$ is not capable of refining style feature maps much and will be less similar to the input one. Therefore we finally apply SMART at resolution $16\times 16$. Another consideration is the exact location since several layers can output style features with the same resolution. In experiments, we find that most detailed attributes are affected by the convolution layer without the function of up-sampling, and therefore we locate SMART after this layer, $F_5$ to be specific. This choice can fulfill high-quality inversion and flexible edits at the same time.

\section{Conclusion and limitation}
Focusing on a learning-based type, we adopt a hierarchical vision Transformer backbone to predict latent codes in $\mathcal{W^+}$ space at token level. By involving feature extraction, latent tokens are capable of disentangling attributes. Then we carefully design a novel SMART block to refine the intermediate style feature maps of the generator in $\mathcal{F}$ space, completing the lost identity information through the cross-attention mechanism. It is also "smart" enough to adapt to the edited cases.
StylePrompter reveals the effectiveness of Transformer for GAN inversion and achieves a balance between reconstruction quality and editing flexibility. However, a limitation in our method is the weakness of inverting out-of-domain details, e.g., characters in the background, since we modulate style features at a shallow layer while the fine details are generally controlled by deeper layers. A possible solution is to stack more SMART blocks, progressively refining the style feature maps to a faithful output.

\begin{acks}
This work was supported by the Natural Science Foundation of China under Grant 62272227.
\end{acks}

\bibliographystyle{ACM-Reference-Format}
\bibliography{reference}

\appendix
\begin{figure*}[htp]
  \centering
  \includegraphics[width=\textwidth]{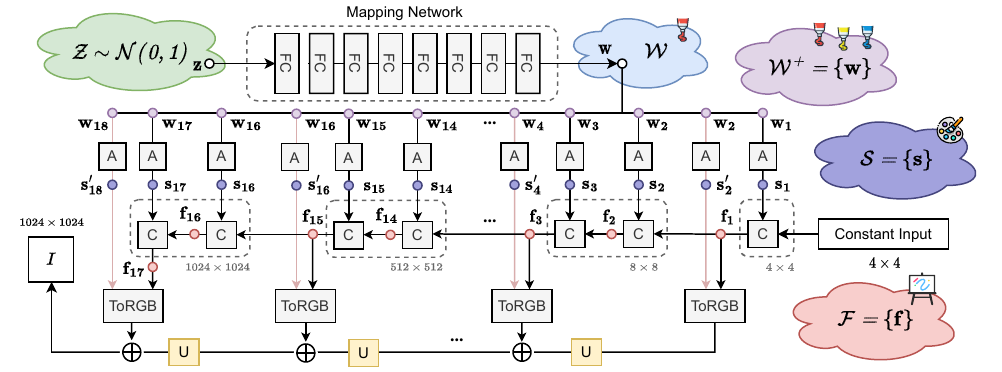}
  \caption{StyleGAN architecture overview. \fbox{A} stands for a learned affine transform, \fbox{C} applies a convolution layer, \fbox{U} represents an up-sample operation respectively. Here exist different latent spaces in the StyleGAN generator, denoted as $\mathcal{Z}$,$\mathcal{W}$, $\mathcal{W^+}$, $\mathcal{S}$, $\mathcal{F}$. As the latent codes $w$ contain information of attributes, we liken $\mathcal{W}$ and $\mathcal{W^+}$ space to \textit{pigments}. Specialized style codes $s$ is like \textit{brush} that will "draw" on the former layer's feature maps. Finally, the most expressive latent space $\mathcal{F}$ can be seen as a \textit{canvas}, where the output features of ToRGB block will be gradually added together.}
  \Description{StyleGAN architecture.}
  \label{spaces}
\end{figure*}
\section{StyleGAN Architecture}
Given latent codes $z\in \mathcal{Z}$ sampled from Gaussian distribution, a non-linear mapping network first produces $w\in \mathcal{W}$. This mapping network is implemented as an 8-layer MLP for the official StyleGAN. The original $\mathcal{W}$ space utilizes the same $w$ codes to control each layer of the synthesis network, i.e., $w_1=w_2=...=w_{18}$, while the extended latent space $\mathcal{W^+}$ involves different $w$ as the input of convolution layers.

In StyleGAN, three convolution layers can be seen as one group whose output tensors have the same resolution but are of different functions. To be specific, the first one will $2 \times$ up-sample the input feature maps, named as \textit{conv\_up}, the second generating more semantic features is named \textit{conv}, the third named \textit{ToRGB} that inverts attributes from feature-level to image-level (dimension change from $C$ to 3 for RGB mode). Each convolution layer will specialize one $w$ to channel-wise style codes $s \in \mathcal{S}$ via a learned affine transformation. These specialized style codes $s$ are actually the convolution kernel weights and will be used to modulate the feature maps $f$ output by the previous non-\textit{ToRGB} convolution layer. 

Notice that the output of $i$-th \textit{ToRGB} will be only used to add upon the previous $(i\mbox{-}1)$-th \textit{ToRGB}'s output, similar to oil painting that continuously paints over a dried base layer, as the illustration in Figure \ref{spaces}. $\mathcal{S}$ and $\mathcal{F}$ space comes out as a collection of styles or feature maps in each convolution layer. Style codes $s$ and feature maps $f$ can have different dimensions. 
\section{StylePrompter Architecture}
\subsection{Details of Backbone}
Swin Transformer is designed with four stages, and each will output down-sampled feature maps, \begin{math}\frac H4 \times \frac H4\times C\end{math}, \begin{math}\frac H8 \times \frac H8\times 2C\end{math}, \begin{math}\frac H{16} \times \frac H{16}\times 4C\end{math}, \begin{math}\frac H{32} \times \frac H{32}\times 8C\end{math} respectively, which will be used in SMART to modulate the intermediate style feature maps of the generator. Each stage has a different depth of attention blocks, controlled by default settings. 

In particular, except for Window-based Multi-head Self-attention (W-MSA), Swin applies a Shifted Window-based Multi-head Self-Attention (SW-MSA) in successive blocks to introduce connection across windows. In our revised latent-involved Swin Transformer, we do not care about what configuration of window partition, but the window count $N$ that latent tokens need to take part in. More specifically, given a feature map with shape $(B, H, W, C)$, where $B$ is the batch size, $H$ and $W$ are patch counts, and $C$ is the dimension. The operation of window partition will reshape the image patches to $(B \times N, \frac{H}{M}, \frac{W}{M}, C)$, where $N$ is the number of windows, $M$ is the window size. Finally changing to patch tokens, it has a shape of $(B \times N, \frac{H}{M} \times \frac{W}{M}, C)$. Our appended latent tokens have an initial shape $(B, T, C)$, where $T$ is the number of latent tokens, e.g., 18 for the generator at $1024 \times 1024$ resolution. Then we replicate it into $(B \times N, T, C)$ in accordance with window count, and then concatenate to each window of partitioned patch tokens as $(B \times N, \frac{H}{M} \times \frac{W}{M} + T, C)$.
This concatenation will be the input of each attention block, encouraging patch tokens and latent tokens to influence each other and learn both image-level features and latent-level information. Both patch tokens and latent tokens apply a residual connection. After self-attention, the latent tokens will separate from patch tokens and turn back to the initial shape through a summation followed by LayerNorm. This design is simple but efficient.

Between stages, a module named Patch Merging will reduce the number of patch tokens (\begin{math}2 \times\end{math} down-sampling of resolution) and enlarge the dimension (\begin{math}2 \times\end{math}), while latent tokens only pass an MLP followed with LayerNorm to match the changed dimension of patch tokens. At the output of the last stage, latent tokens keep the same number $T$ as the input, but the dimension goes up to 8C; the number of patch tokens is reduced to \begin{math}\frac H{32} \times \frac H{32}\end{math} and the dimension reaches up to 8C. Finally, the prediction head implemented as a 3-layer MLP with Tanh activation in between will transfer the latent tokens to latent codes in $\mathcal{W^+}$. 

Existing inversion methods commonly apply numerous convolution layers to predict $\mathcal{W^+}$ latent codes through the extracted feature maps of the encoder, and we instead embed latent codes as tokens, which is not only super effective and low-cost in time and scale but also aware of recognizing attributes, benefiting to disentanglement.
\subsection{Details of SMART}
\begin{figure}
  \centering
  \includegraphics[width=\linewidth]{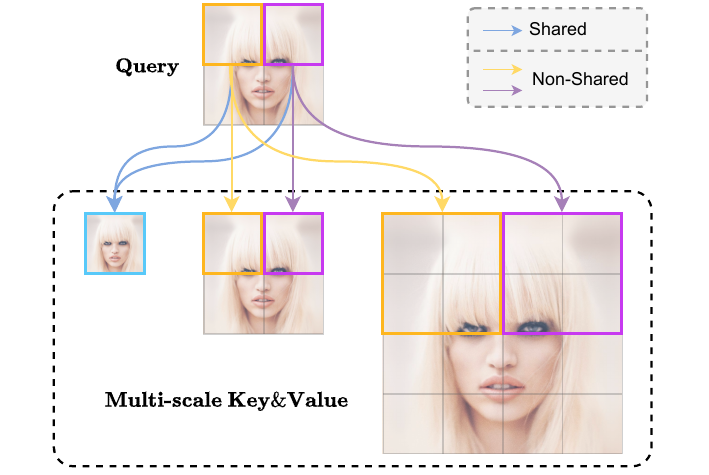}
  \caption{Details of Multi-scale Attention. Key and value elements are linear projections from the encoder's feature maps which have different sizes. Queries projected from the style feature maps of the generator will search for certain spatial indexes of keys and values according to the ratio of resolutions. For example, if the feature maps of key and value have a smaller size than the query, several query elements will search the same element of key and value, denoted as \textit{shared} in the blue box.}
  \Description{Details of Multi-scale Attention.}
  \label{multi-scale_attention}
\end{figure}
Given the style feature maps $F_l$ at the $l$-th convolution layer of the generator, and the multi-scale image feature maps $\{P_s\}_{s=1}^S$ extracted by the Swin backbone, where $S$ is the number of stages, we then calculate the Q, K, V by learned linear projections:
\begin{displaymath}
    Q=F_lW^Q + b^Q, K=\{P_sW^K_s+b^K_s\}^S_{s=1}, V=\{P_sW^V_s+b^V_s\}^S_{s=1}
\end{displaymath}
$K$ and $V$ are concatenations of the multi-scale feature maps, which means only one attention between $F_l$ and $\{P_s\}^S_{s=1}$ is calculated.

Local attention is also employed and each query will search for specific key and value elements corresponding to the same spatial location, which is computationally cheap. As shown in Figure \ref{multi-scale_attention}, if $s$-th feature maps have an equal shape with $F_l$, queries seek key and value elements in the same spatial location. 
Other cases will depend on a calculated ratio \begin{math}\{R_s\}^S_{s=1}={H_{F_l}}/{H_{P_s}}\end{math}. 
For smaller scales, \begin{math}R\times R\end{math} neighbors in queries may search the same key and value elements (\textit{shared} in blue line). For larger scales, which are supposed to have much more low-level image details, each query will search \begin{math}\frac{1}{R}\times \frac{1}{R}\end{math} ($R\textless 1$) keys and values elements to obtain adequate information for reconstruction. 
Recall that we skip the softmax and scaled operation when calculating the production of $Q$ and $K$. Our designed cross-attention in SMART aims to quantify the lost identity, which is different from the standard cross-attention module. 

The output of the cross-attention block will be added to $F_l$ as a residual. After another residual connection of FFN, the refined style feature maps will be fed to the next convolution layer of the generator to complete inversion. It is notable that we apply additional weight $\beta$ at both residual connections for flexible editing.

\section{Experimental Supplement}
\subsection{Implementation Details}
All experiments are implemented on a single NVIDIA GeForce RTX 3090. 
When computing the identity loss, we follow E2Style \cite{wei2022e2style} to employ a multi-layer identity loss that calculates the cosine similarity between the multi-layer features of the image pair. The hyper-parameters are set to $\lambda_1=1.0$, $\lambda_2=0.6$, $\lambda_3=0.1$, $\lambda_4=0.1$ for human facial domain. For SMART we set $\lambda_4=0$, and other parameters are the same. 

\begin{figure*}
  \centering
  \includegraphics[width=\linewidth]{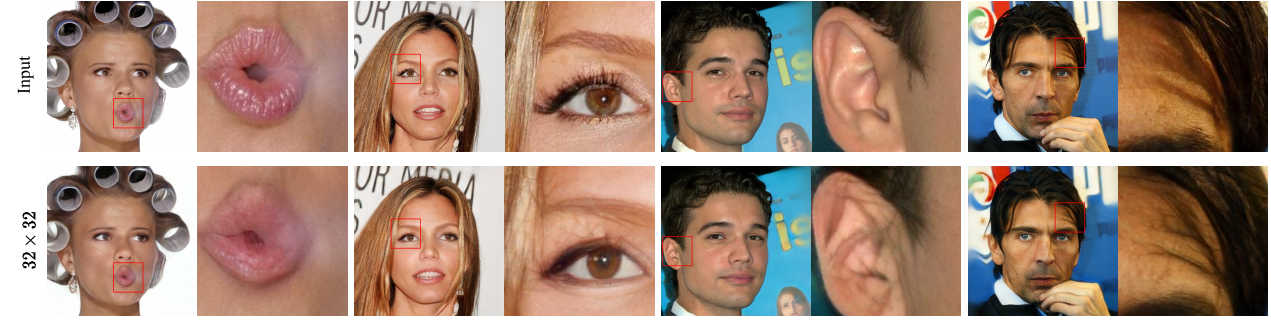}
  \caption{SMART at resolution $32\times 32$. Although it is capable of inverting out-of-domain details such as the letters in the background, zoom-in patches show that the inversion results are not realistic, appearing unreal lifelike textures.}
  \label{SMART_higher_resolution}
\end{figure*}

\begin{table}
  \caption{Quantitative comparison for SMART at different resolutions. SMART at the highest resolution gets the best scores in all metrics, but the inversion results are not desirable for human perceptual preference.}
  \label{SMART_resolution_metrics}
  \begin{tabular}{cccc}
    \toprule
    Resolution & $L_2$ & LPIPS & MANIQA \\
    \midrule
    $8 \times 8$ & 0.033 & 0.1353 & \textbf{0.0188} \\
    $16 \times 16$ & \underline{0.022} & \underline{0.0894} & 0.0175 \\
    $32 \times 32$ & \textbf{0.013} & \textbf{0.0486} & \underline{0.0179} \\
    \bottomrule
\end{tabular}
\end{table}

\subsection{Metric Discussion}
As we define \textit{quality} as a combination of \textit{} and \textit{realism}, it is necessary to estimate both of them. fidelity, also called \textit{distortion} or \textit{faithful} in previous works, should be computed between the image pair of input and the inverted one. Full reference metrics such as the commonly used $L_2$, LPIPS can measure fidelity.

However, realism presents a visual preference of human beings, which has been studied as \textit{perceptual quality} in e4e. Although image quality assessment (IQA) and image aesthetic quality assessment (IAQA) have become new and fascinating areas of research in recent years, existing methods focus on distinguishing the degraded images, while not taking realism into consideration. These unreal
lifelike textures, especially in the interested region such as the facial features for the human face domain and the fur for the animal domain, are unsatisfying for the human eyes’ perception, but still a challenge for deep-learning quality models \cite{yang2022maniqa}. Despite MANIQA we used in this paper to estimate realism, it is not good enough for recognizing unreal textures and needs domain-specific knowledge. As listed in Table \ref{SMART_resolution_metrics}, SMART at resolution $32 \times 32$ gets higher MANIQA scores than that of resolution $16 \times 16$, but we do emphasize that the inverted images by this model possess undesirable textures, as shown in Figure \ref{SMART_higher_resolution}. 
It reveals that existing models in image quality assessment are not able to discriminate between real and unreal features, and thus choosing an appropriate quality metric is still worthy of investigation. 

It is our hope that deep-learning models can reflect human perceptual preferences in the future.

\subsection{Disentanglement of $\mathcal{W^+}$}
\begin{figure}
  \centering
  \includegraphics[width=\linewidth]{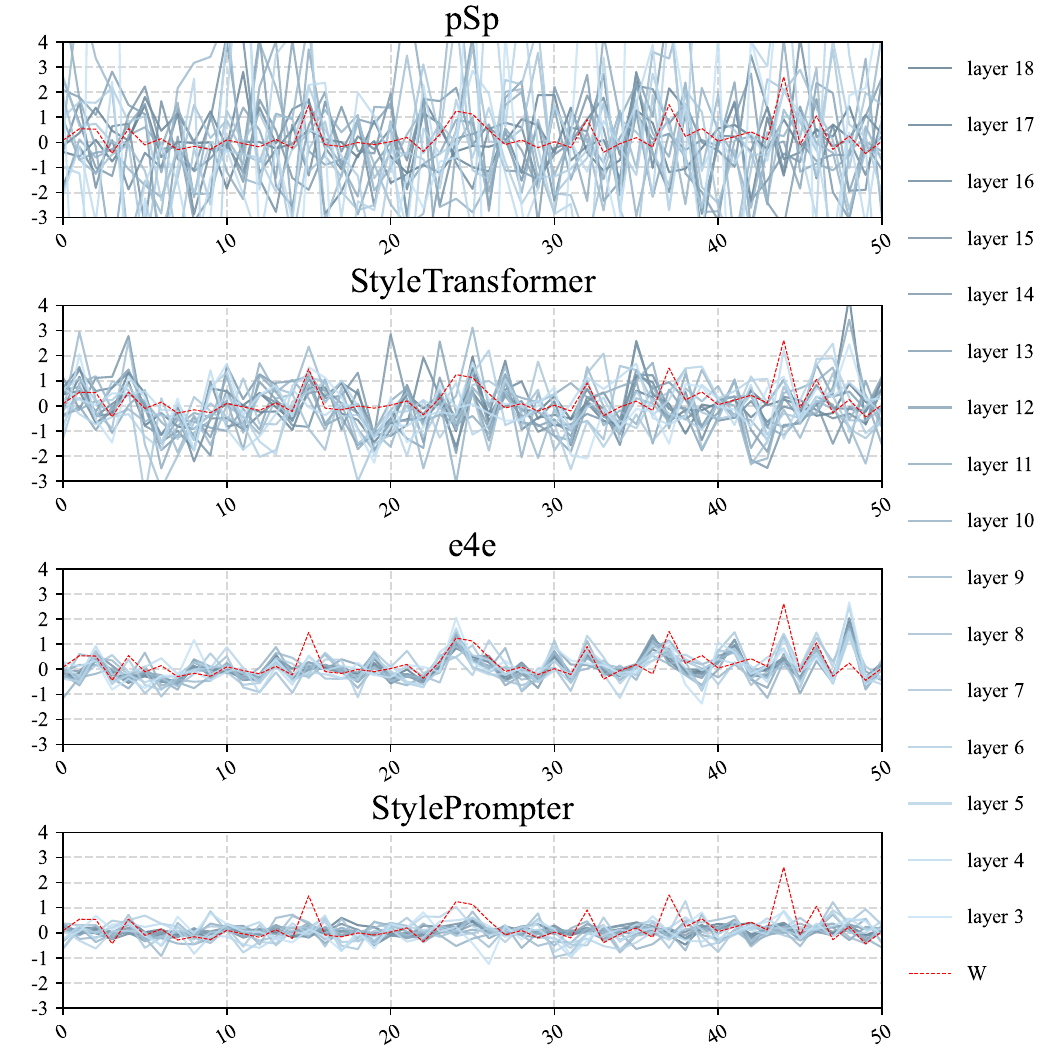}
  \caption{Disentanglement comparison for $\mathcal{W^+}$-based methods, we plot the first 50 dimensions of one sample's inverted latent codes by pSp, e4e, StyleTransformer and StylePrompter (without SMART) to study the correlation between layers. pSp and StyleTransformer produce turbulent latent codes, causing unreal textures in the inverted images.}
  \Description{Respond example 1.}
  \label{disentanglement_per_layer}
\end{figure}

The comparison of different $\mathcal{W^+}$-based methods with the original $\mathcal{W}$ space in Figure \ref{disentanglement_per_layer} demonstrates that pSp, StyleTransformer produce turbulent latent codes among layers, which can cause unreal textures in the inverted images. Most layers in e4e are in line with each other, indicating entanglement. In contrast, StylePrompter is capable of locating different values at different layers. It may produce positive values in some layers, but negative values in other layers at the same dimension.

We further utilize style mixing to explore what it will affect visually. 
Let $w^s$ represent the latent codes of the source image, $w^r$ as the reference image. For StyleGAN in the face domain with 18 tokens, we conduct \textit{progressively replacing}, \textit{one-layer exchanging} and \textit{interpolation} as follows:
\begin{align}
    & w_{gr} = \{w^r_1, ..., w^r_{k}, w^s_{k+1}, ... w^s_{18}\}, &k=1,...,18 \nonumber \\ 
    & w_{oe} = \{w^s_1, ..., w^s_{k-1}, w^r_k, w^s_{k+1}, ... w^s_{18}\}, &k=1,...,18 \nonumber \\
    & w_{inter} = (1 - \sigma) \cdot w_r \times \sigma \cdot w_s, &\sigma=0,...,1 \nonumber 
\end{align}

\begin{figure*}[h]
  \centering
    \subfigure[One-layer Exchanging]{
      \includegraphics[width=\textwidth]{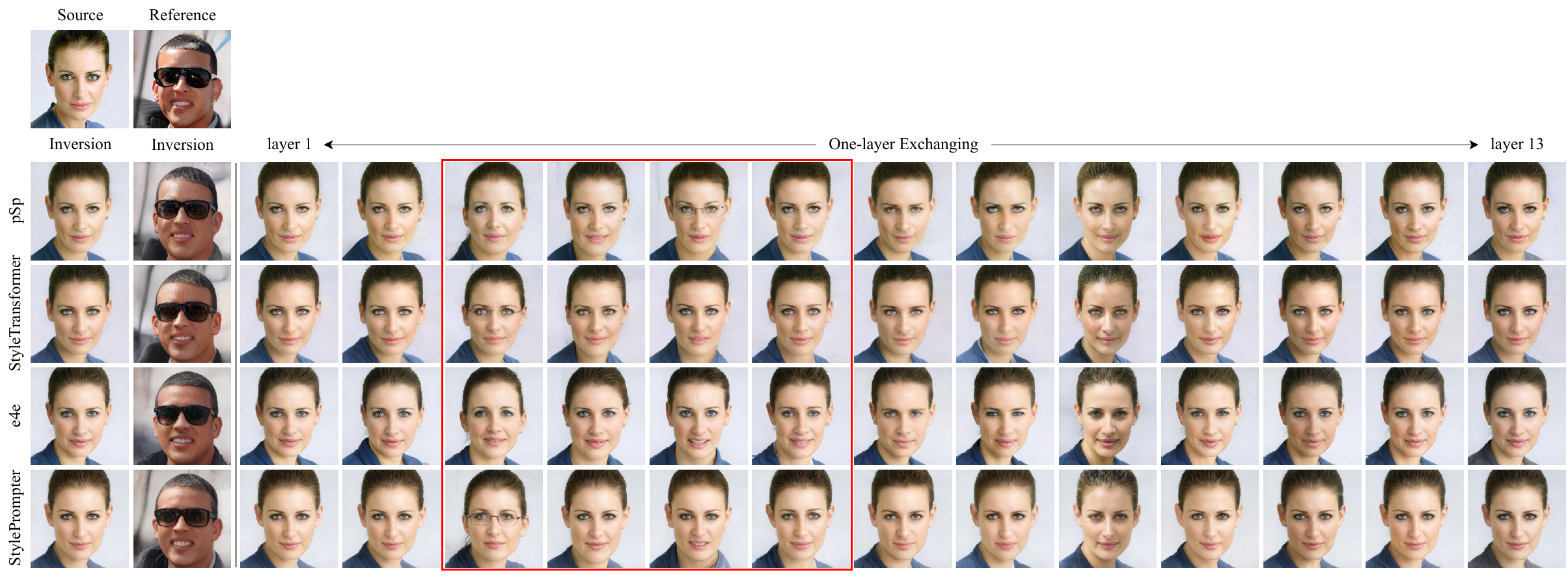}
    }
    \quad
    \subfigure[Interpolation]{
      \includegraphics[scale=0.46]{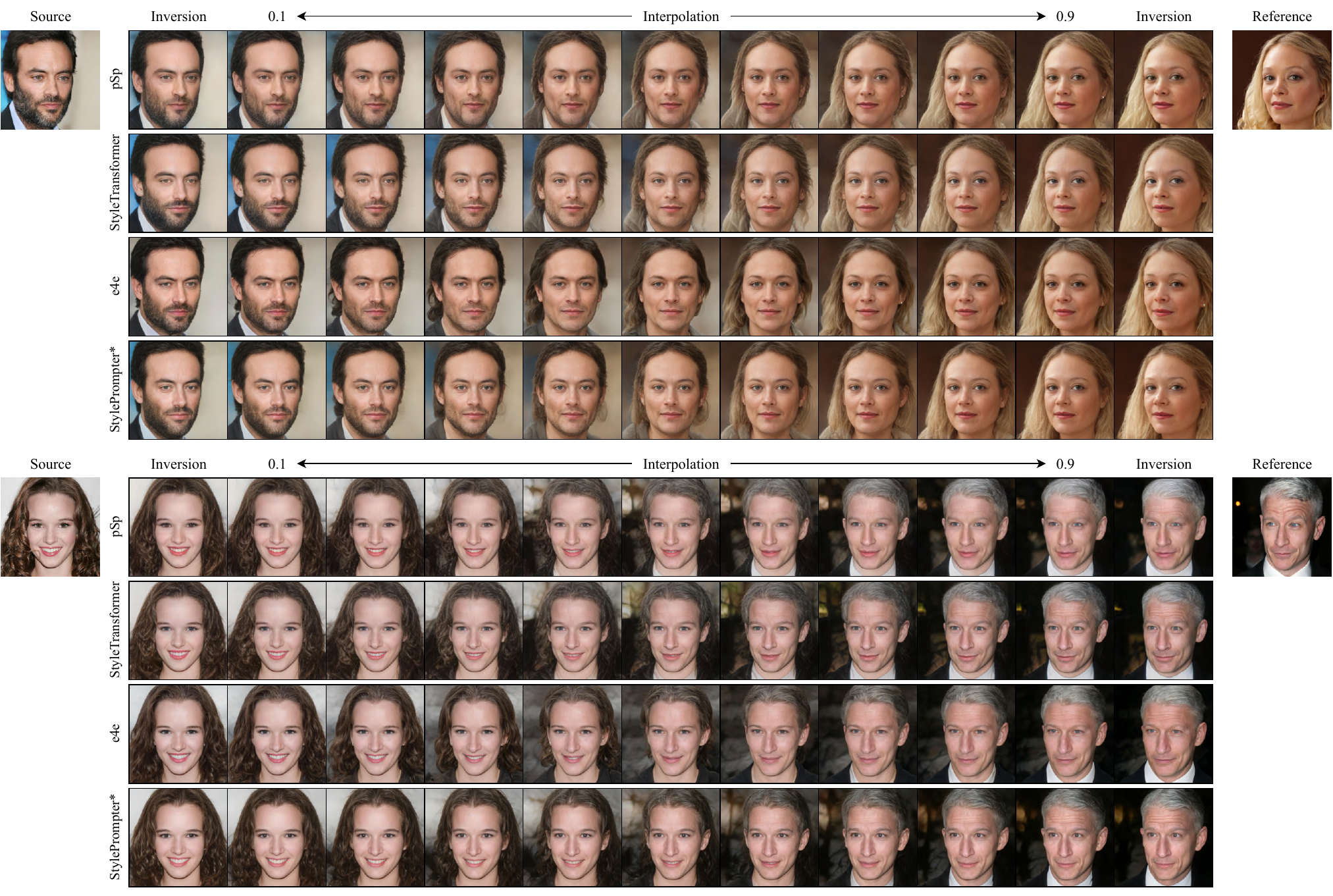}
    }      
  \caption{We compare different $\mathcal{W^+}$-based methods by style mixing. (a) One-layer exchanging comparison. We replace one of the source image's latent codes with that of the reference image. In the red box, it indicates that StylePrompter is capable of locating more value at the layer that gives the most active response to certain attributes. For example, the eyeglasses attribute in the latent codes predicted by StylePrompter mostly concentrates at layer 3, while other methods may disperse into several layers. As with the comparison of progressively replacing, StylePrompter produces the cleanest and sharpest changes in one-layer exchanging. (b) Interpolation comparison. As a benefit of disentanglement, StylePrompter is capable of producing smoother intermediate interpolation images than other methods.}
  \Description{Style mixing comparison.}
  \label{style_mixing_oe_it}
\end{figure*}

\begin{figure*}[h]
  \centering
    \subfigure[Age]{
      \includegraphics[scale=0.6]{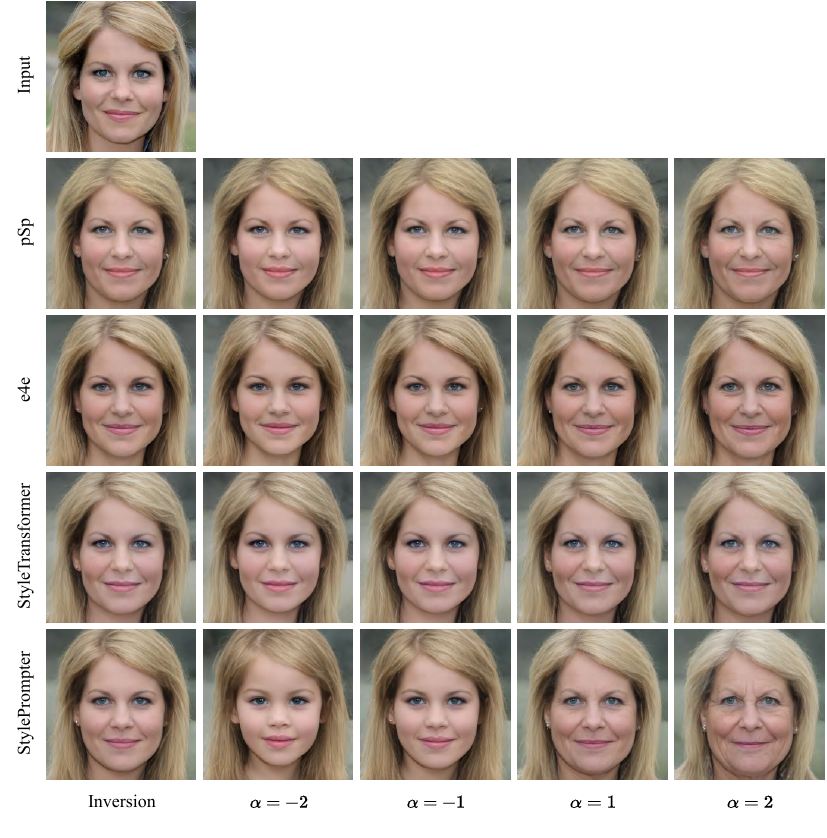}
    }
    \subfigure[Pose]{
      \includegraphics[scale=0.6]{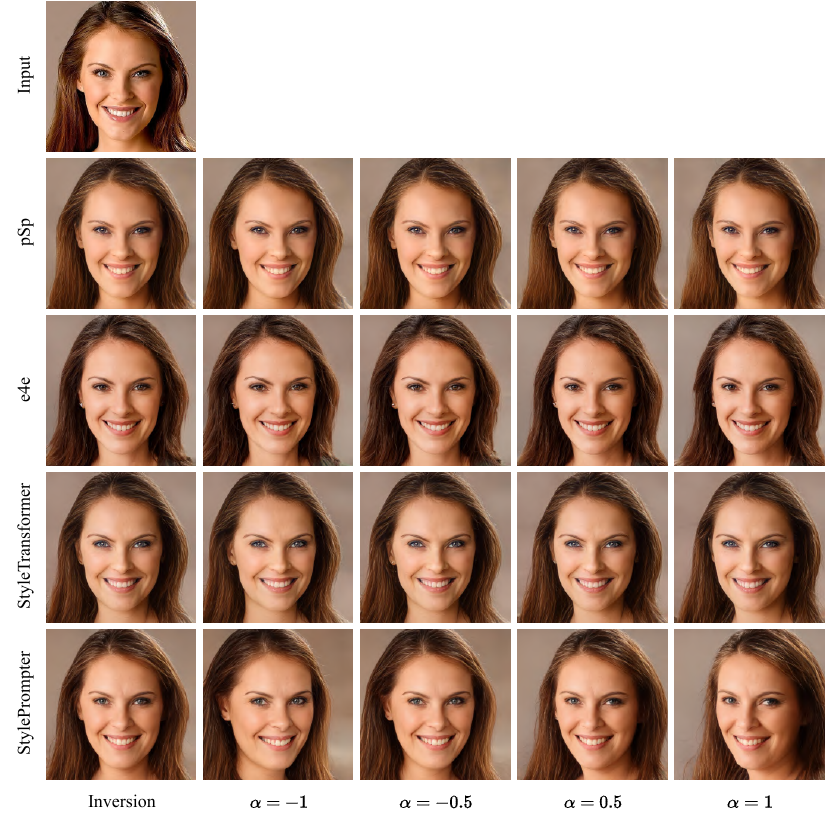}
    }
    \quad
    \subfigure[Narrow Eyes]{
      \includegraphics[scale=0.6]{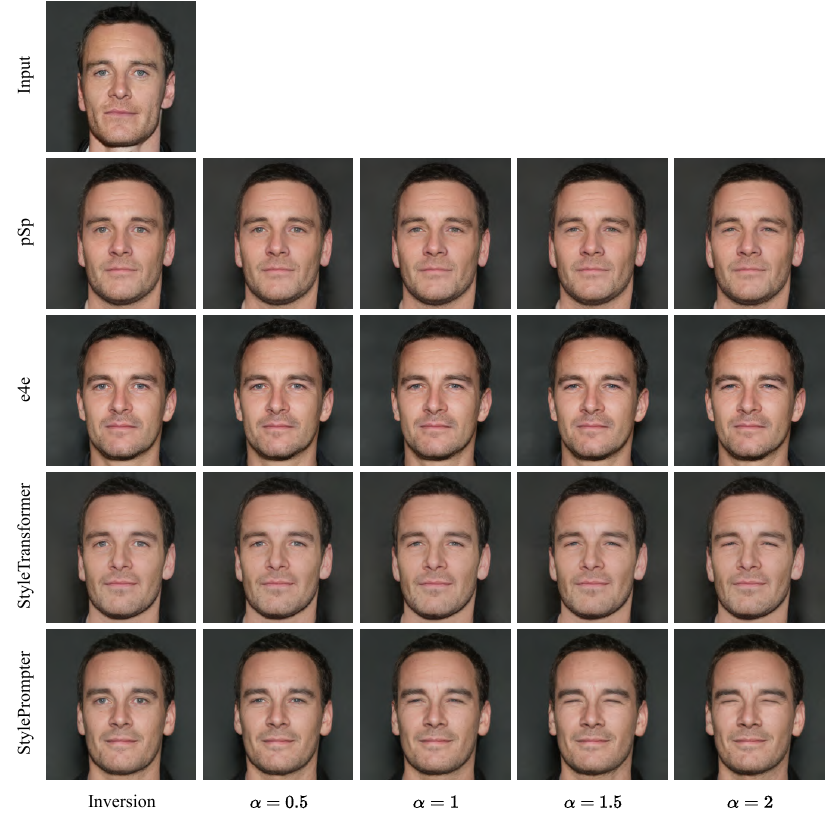}
    }
    \subfigure[Beard]{
      \includegraphics[scale=0.6]{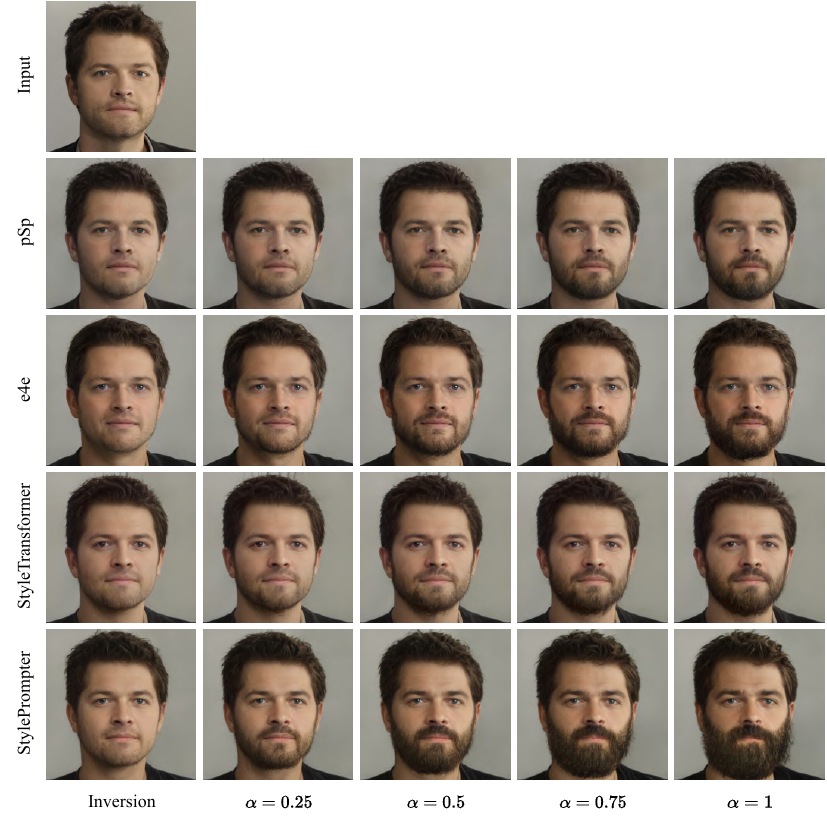}
    }     
  \caption{We compare the manipulation results among $\mathcal{W^+}$-based methods under the same magnitude $\alpha$. Each attribute that is edited is indicated below the image. Each column shows the edited image under the same $\alpha$ of specific attribute direction by different methods. In every case, StylePrompter is able to provide results that have a greater variation than other methods.}
  \Description{Alpha comparison.}
  \label{disentanglement_alpha}
\end{figure*}

The additional comparisons of one-layer exchanging and interpolation are shown in Figure \ref{style_mixing_oe_it}. To further show our more disentangled $\mathcal{W^+}$ space, we compare the manipulation results among $\mathcal{W^+}$-based methods under the same magnitude $\alpha$ in Figure \ref{disentanglement_alpha}.

\subsection{Controllability of SMART}
\begin{figure}
  \centering
  \includegraphics[width=\linewidth]{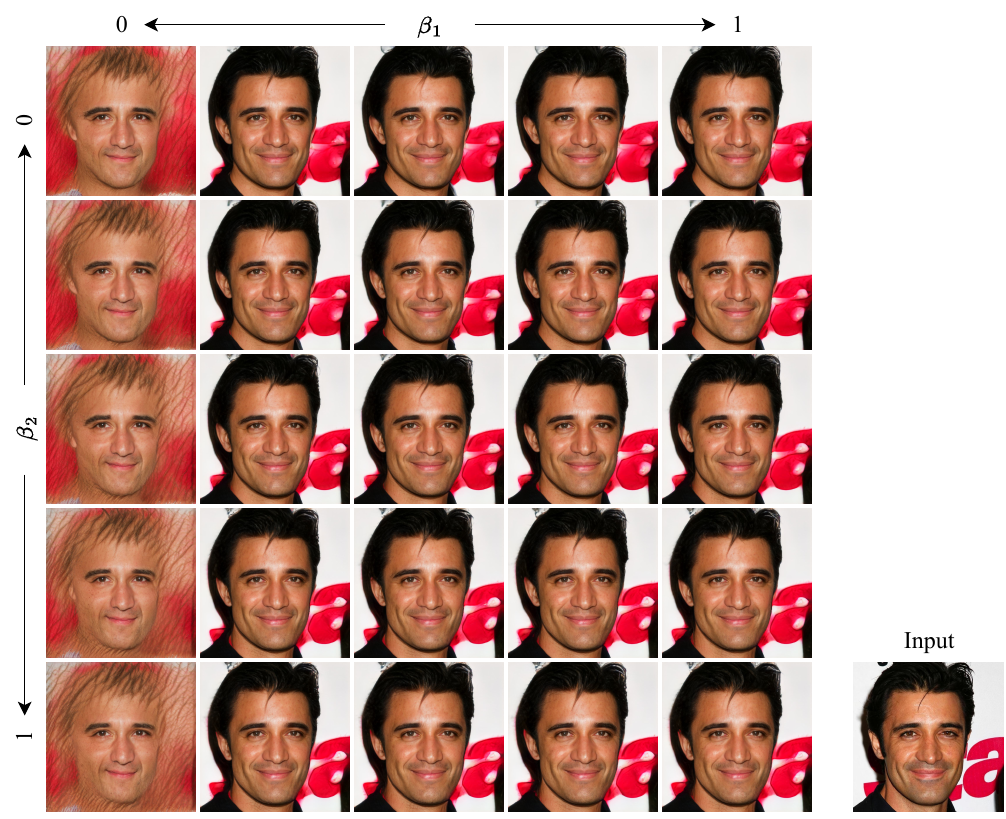}
  \caption{SMART with LayerNorm will hurt the controllability of $\beta$ that the change of manipulated images is not interpretable and explainable.}
  \Description{SMART with norm layer.}
  \label{beta_with_norm}
\end{figure}
\begin{figure}
  \centering
  \includegraphics[width=\linewidth]{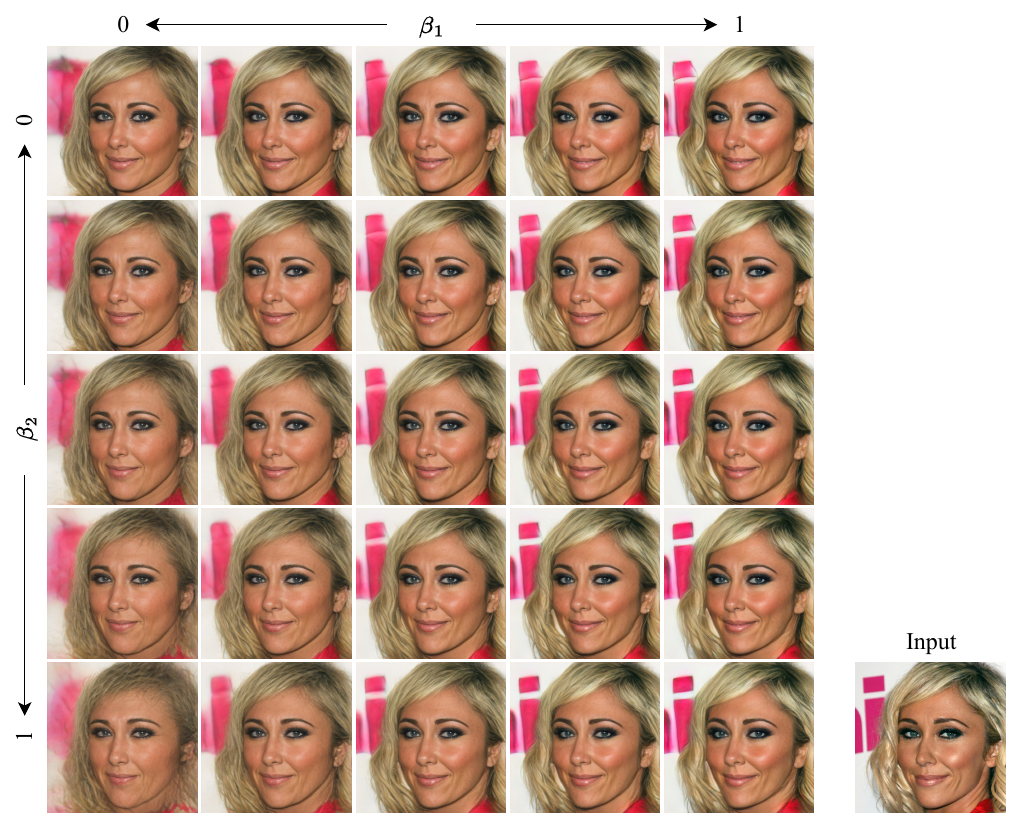}
  \caption{Another example of our designed controllable SMART. The vertical changing demonstrates FFN learns to suppress invalid and unimportant values that hair and background will disappear as $\beta_2$ increases. The horizontal change indicates that cross-attention can retrieve lost identity information from the encoder's feature maps. The larger the value of $\beta_1$, the more similar it is to the input image.}
  \Description{SMART controlled by beta.}
  \label{beta_example2}
\end{figure}
\begin{figure}
  \centering
  \includegraphics[width=\linewidth]{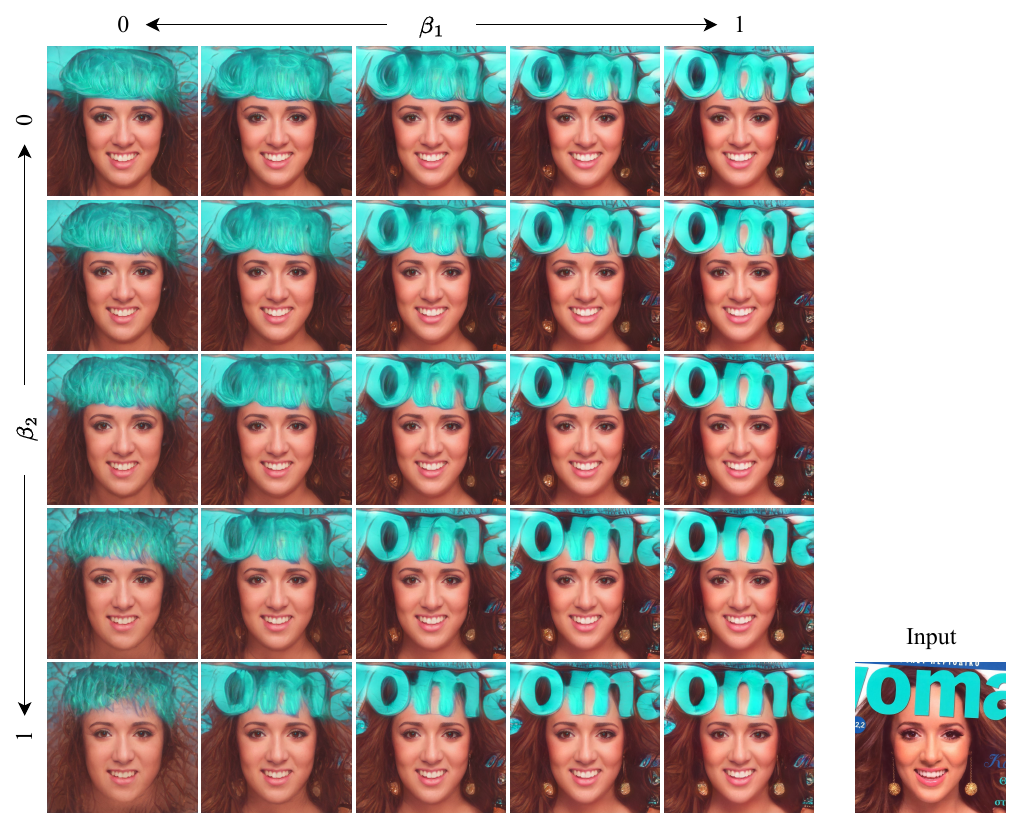}
  \caption{Another example of our designed controllable SMART. The changing of fluorescent characters sheds light on how SMART works.}
  \Description{SMART controlled by beta.}
  \label{beta_example3}
\end{figure}
\begin{figure}
  \centering
  \includegraphics[width=\linewidth]{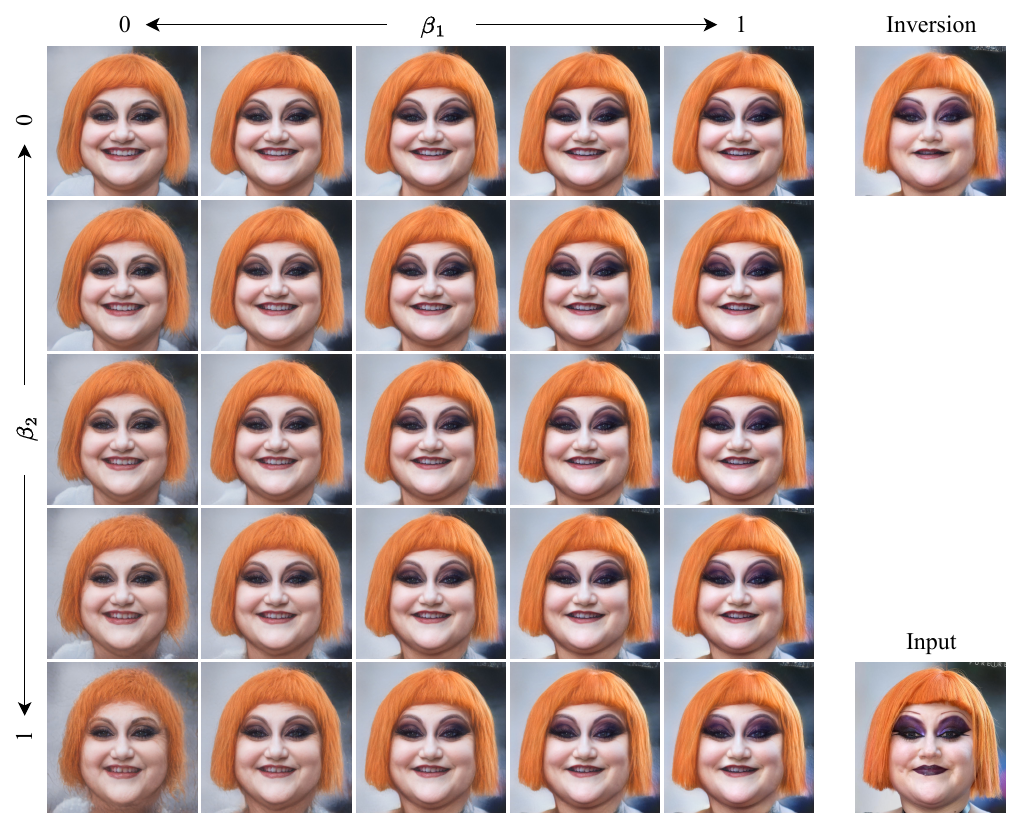}
  \caption{Controllability of SMART for editing task. The increased $\beta_1$ can retrieve more identity information but weaken the smiling strength. The change caused by $\beta_2$ is subtle. As FFN focuses on refining invalid values, the edited attributes can also be suppressed, resulting in smiling fades.}
  \Description{SMART controlled by beta.}
  \label{beta_example4}
\end{figure}
\begin{figure}
  \centering
  \includegraphics[width=\linewidth]{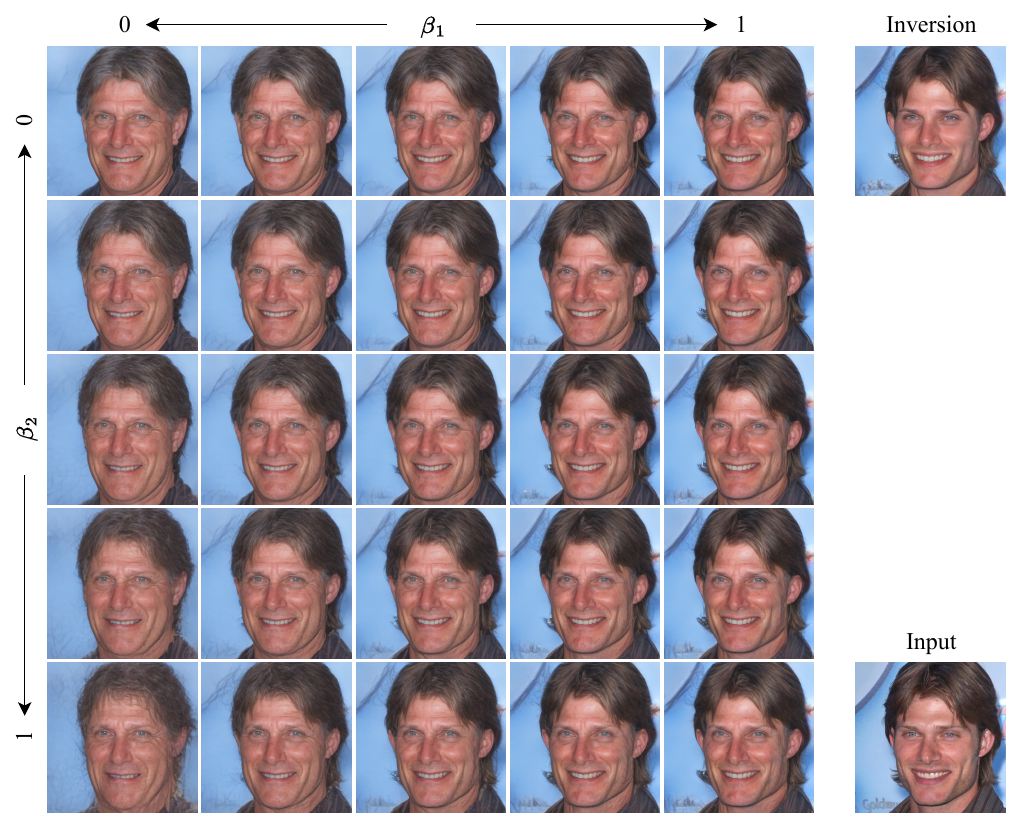}
  \caption{Another example of controllable SMART for age manipulation. In this case, identity consistency is not necessary, thus we can use smaller $\beta_1$ to change appearance. Another observation is that FFN can influence color, e.g., hair will darken if $\beta_2$ increased.}
  \Description{SMART controlled by beta.}
  \label{beta_example5}
\end{figure}

As mentioned above, we append two additional weights $\beta_1$ and $\beta_2$ to provide Style-driven Multi-scale Adaptive Refinement Transformer (SMART) with controllability, while involved with LayerNorm can hurt this ability. This ablation study is shown in Figure \ref{beta_with_norm}. We also provide additional examples of controllable SMART in Figure \ref{beta_example2}, \ref{beta_example3}, \ref{beta_example4} and \ref{beta_example5}, the first two for inversion and the other two with respect to editing cases.

\subsection{Pose Manipulation}
\begin{figure}
  \centering
  \includegraphics[width=\linewidth]{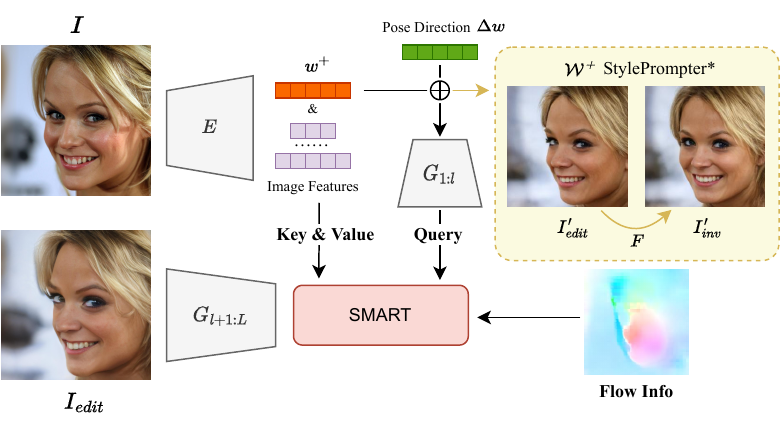}
  \caption{A framework for manipulating pose. We involve an optical flow model to estimate the per-pixel motion between the inversion-edit image pair via StylePrompter* in $\mathcal{W^+}$. The predicted flow information will update the location of key and value elements that queries need to retrieve from. The output edited image $I_{edit}$ can maintain identity consistency with the desired pose, compared with $I_{edit}'$.}
  \Description{pose manipulation.}
  \label{pose_manipulation}
\end{figure}

Owe to the local attention in SMART, we build a flow-involved framework to manipulate the pose. As illustrated in Figure \ref{pose_manipulation}, it consists of StylePrompter and an optical flow model, e.g., RAFT \cite{teed2020raft}.

Let $I$ denotes the input image, we first obtain its corresponding latent codes $w^+ \in \mathcal{W^+}$ and image features from the Swin Transformer backbone and the prediction head, together as $E$. Theoretically, the pose variation will not hurt the identity consistency but will change spatial context only. Therefore, we involve an optical flow network to predict the motion of pixels. More specifically, we manipulate the inverted latent codes $w^+$ by adding a prior direction $\Delta w$ associated with pose to get an inversion-edit image pair in $\mathcal{W^+}$, which is supposed to have the same motion of pixels with the pair in $\mathcal{F}$ space:
\begin{displaymath}
    \Delta^x, \Delta^y = F(I_{inv}', I_{edit}') \approx F(I_{inv}, I_{edit})
\end{displaymath}
where $F$ is the flow model, $\Delta^x$ and $\Delta^y$ are the horizontal and vertical flow estimated by $F$, we produce the inversion and edited images $I_{inv}'=G(E(I))$ and $I_{edit}'=(G(E(I)+\alpha \cot \Delta w))$ via StylePrompter*, where $G$ is the original generator of StyleGAN without SMART. To involve $\mathcal{F}$, the offset of pixels between the image pair of $I_{inv}$ and $I_{edit}$ which are the inversion and edited images via the full size of StylePrompter are required, while the target image $I_{edit}$ is unknown. Based on our hypothesis, the motion between $I_{inv}'$ and $I_{edit}'$ can approximate the desired pixels offsets. We then utilize the flow information to update the index of key and value elements that the query elements will retrieve. For example, in the case of inversion, each query element $q_{i,j}$ ($i\in\{1,2,...,H\}$ and $j\in\{1,2,...,W\}$, where $H$ and $W$ are height and width of the style feature map) will search key and value elements in the corresponding index $k^{q_{i,j}}_s, v^{q_{i,j}}_s$, where $s\in\{1,2,...,S\}$, $S$ is the number of stages. While in the case of pose manipulation, the per-pixel motion estimated by the optical flow model will constrain the query in spatial location $(i, j)$ to search the key and value elements with respect to the query in spatial location $(i+\Delta^x_i, j+\Delta^y_j)$. Finally, the target image $I_{edit}$ can be obtained, with a changed pose and preserved identity, compared with $I_{edit}'$. Simply, we formulate this flow-involved pose manipulation as follows:
\begin{displaymath}
    I_{edit} = SMART(F_l,\{P_s\}^S_{s=1}, \Delta)
\end{displaymath}
where $\Delta = (\Delta^x, \Delta^y)$. 

\subsection{Other Results}
\begin{figure*}[h]
  \centering
  \includegraphics[width=\textwidth]{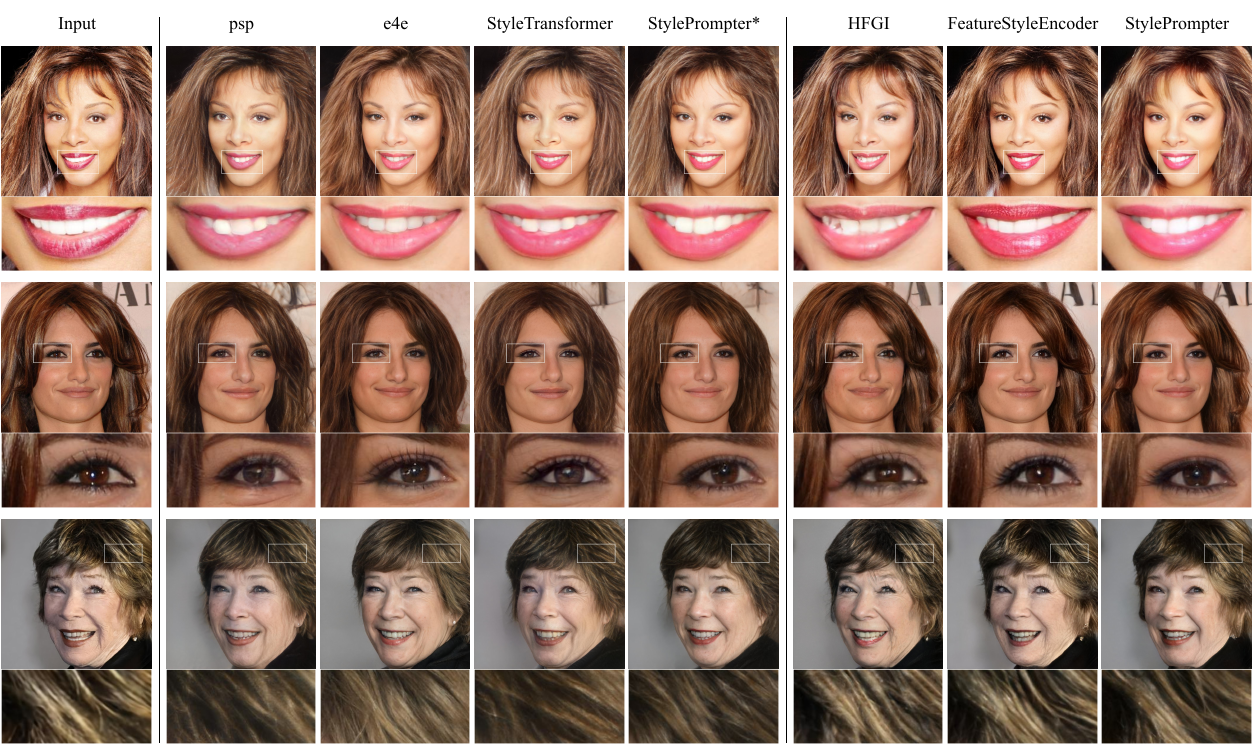}
  \caption{Inversion results and zoom-in patches via different inversion methods. Among $\mathcal{W^+}$-based methods, e4e produces the most realistic images, due to the approximation of $\mathcal{W}$ space. Also close to $\mathcal{W}$, the inverted images of StylePrompter have fewer artifacts than pSp and StyleTransformer. Among $\mathcal{F}$-involved methods, fake textures will appear in the inverted images of HFGI especially for out-of-domain facial features. The reason is that HFGI modifies style features based on the difference between the input image and coarse inverted image, while excessive misalignment will influence the capability of refinement.}
  \Description{Realism Comparison}
  \label{realism_comparison}
\end{figure*}

\begin{figure*}[h]
  \centering
  \includegraphics[scale=0.95]{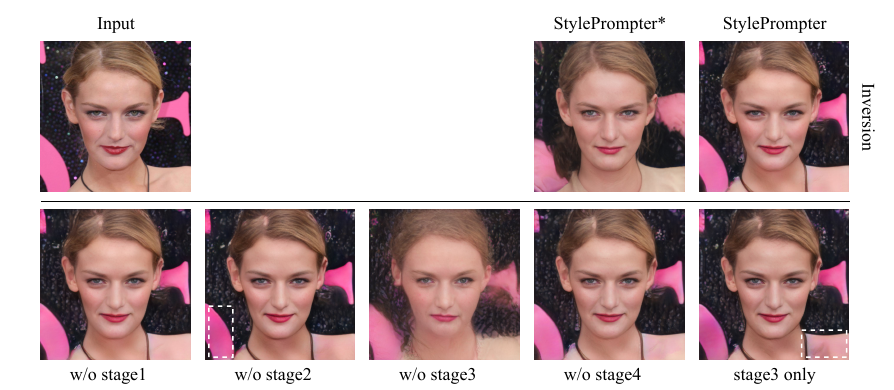}
  \caption{We remove one of the stages from key and value elements in SMART to investigate the influence of multi-scale features. The results suggest that most information concentrates on stage 3, which has the most depth. Although the effects of other stages are subtle, we do emphasize that the identity information in stage 3 is not adequate enough to reconstruct a faithful image, comparing the inversion via the full model of StylePrompter and the inverted image using stage 3 only.}
  \Description{Ablation of multiple stages.}
  \label{ablation_stage}
\end{figure*}

\begin{figure*}[h]
  \centering
  \includegraphics[width=\textwidth]{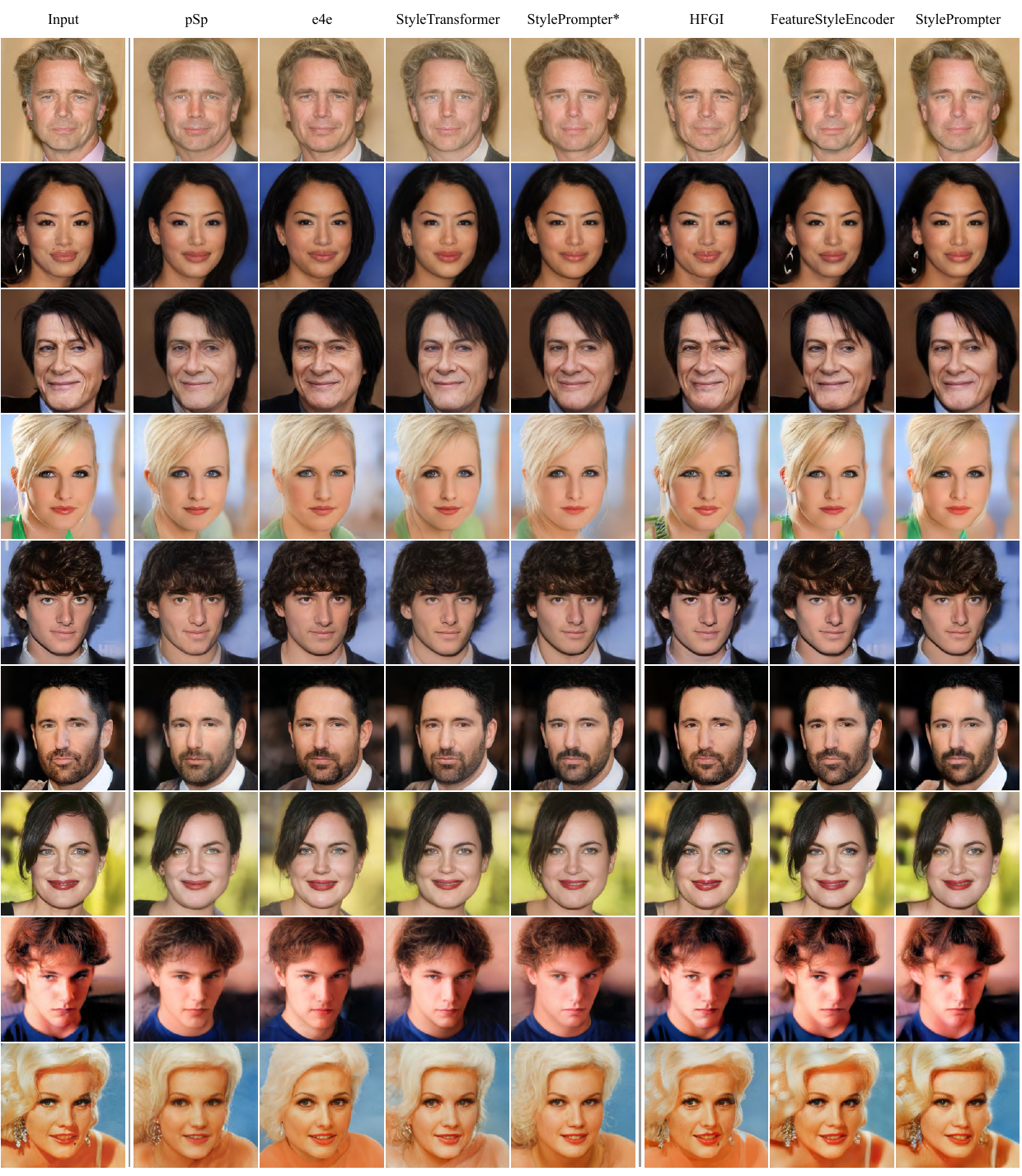}
  \caption{Inversion results of simple instances via different encoder-based GAN inversion methods. The background of input images is a solid color or simply blurred.}
  \Description{Inversion results simple.}
  \label{inversion_results_simple}
\end{figure*}

\begin{figure*}[h]
  \centering
  \includegraphics[width=\textwidth]{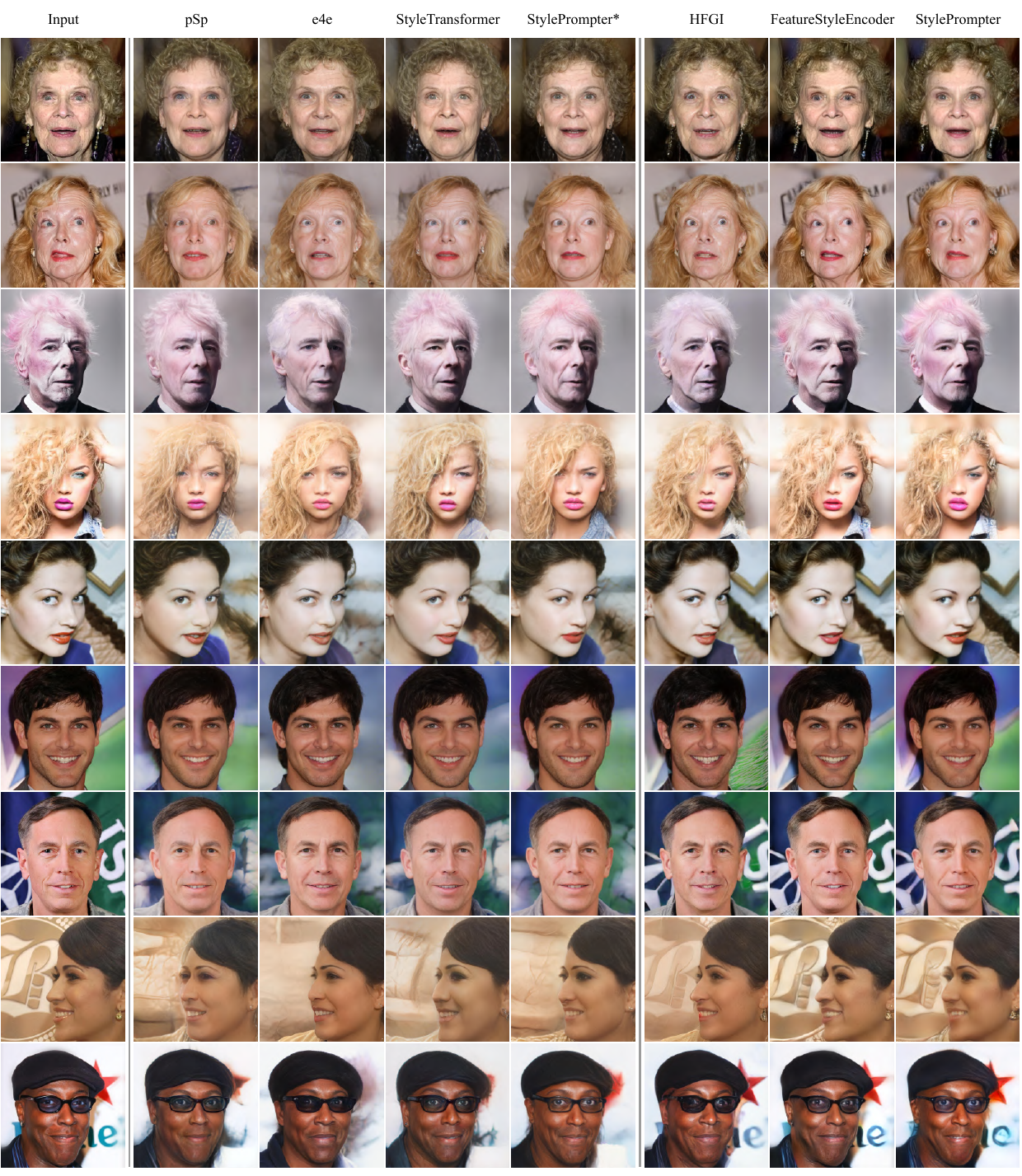}
  \caption{Inversion results of hard instances via different encoder-based GAN inversion methods. These samples have out-of-domain patterns such as hairstyle, exaggerated facial expressions, make-up, or letters in the background.}
  \Description{Inversion results hard.}
  \label{inversion_results_hard}
\end{figure*}

\begin{figure*}[h]
  \centering
  \includegraphics[width=\textwidth]{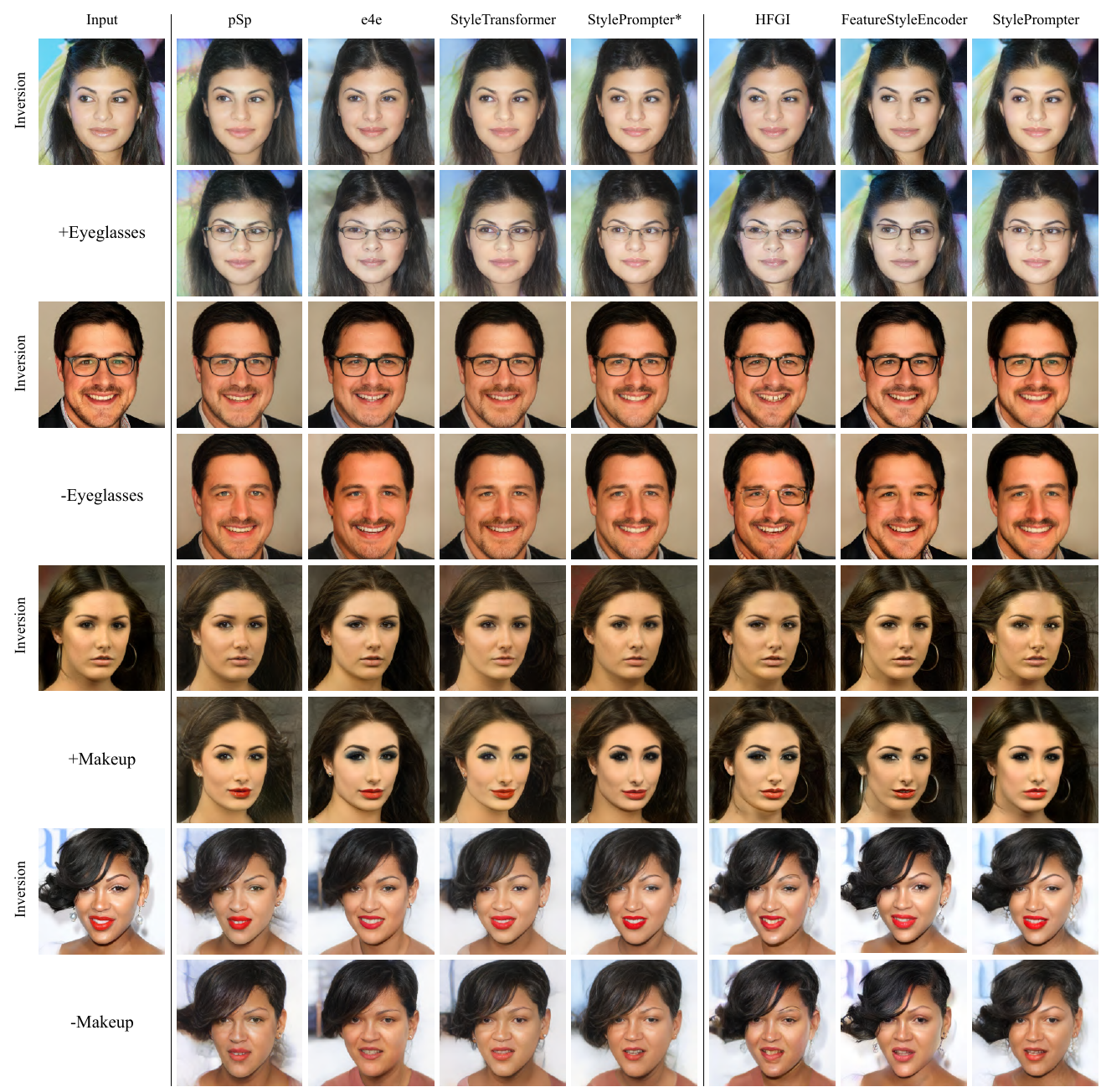}
  \caption{Editing comparison for $\mathcal{W^+}$-based methods (columns 2-5) and $\mathcal{F}$-involved methods (columns 6-8). Notice that the direction magnitude $\alpha$ may differ between methods. For each attribute, we employ "add" and "minus" and control the editing results visual-oriented toward the same degree of target attributes. With better editability, $\mathcal{W^+}$-based methods can produce edited images with desirable attributes, but not faithful enough. HFGI and FeatureStyleEncoder refining style features in inappropriate manners ruin the editability of $\mathcal{F}$, and cannot deal with edit cases, especially for "minus" attributes.}
  \Description{Additional editing results.}
  \label{editing_comparison_additional1}
\end{figure*}

\begin{figure*}[h]
  \centering
  \includegraphics[width=\textwidth]{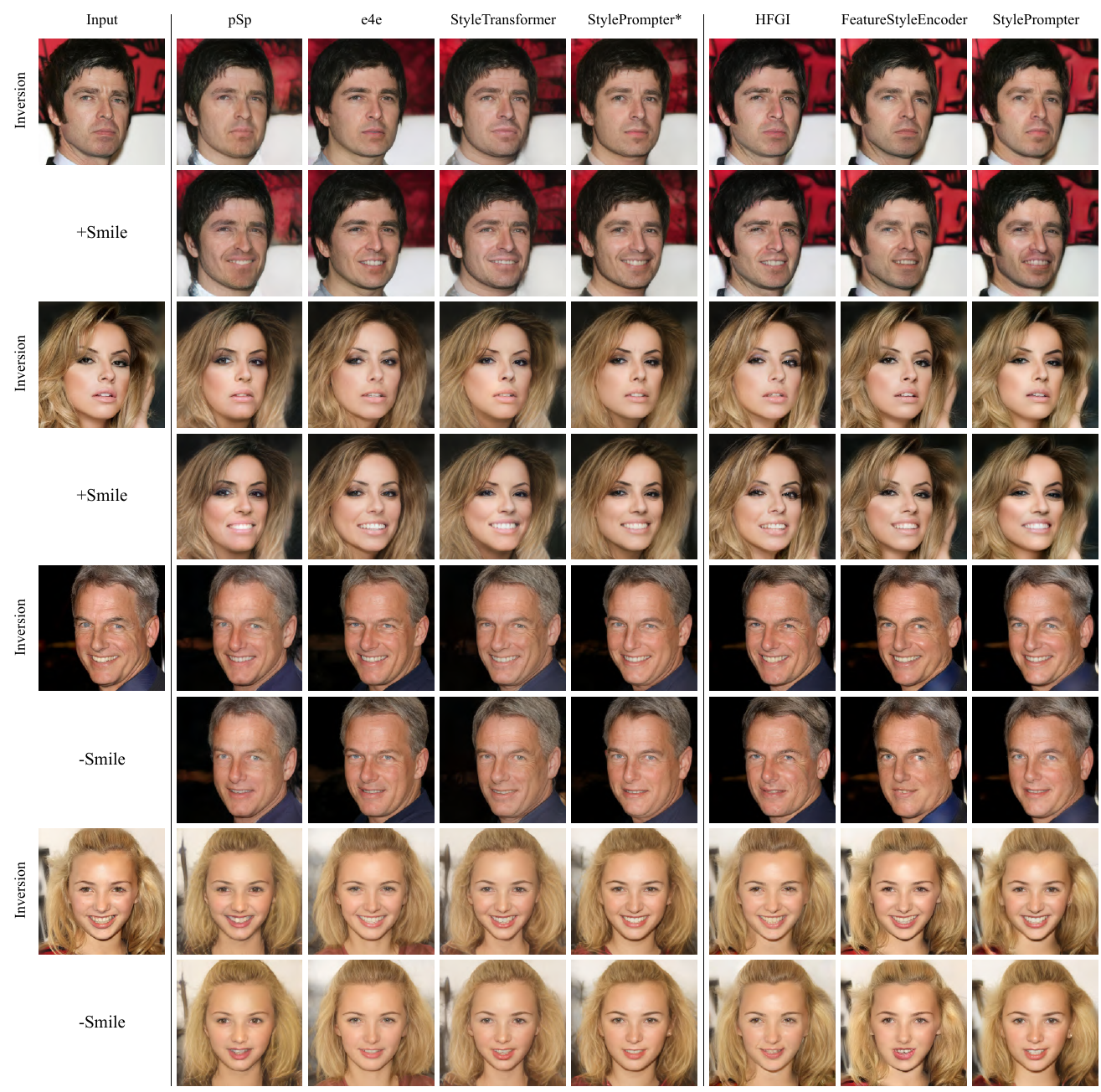}
  \caption{Editing comparison for $\mathcal{W^+}$-based methods (columns 2-5) and $\mathcal{F}$-involved methods (columns 6-8). Notice that the direction magnitude $\alpha$ may differ between methods. We do emphasize that \textit{smiling} is different from \textit{mouth open} and has more semantic meaning. More specifically, opening the mouth only influences the area near the mouth, but a smile may cause a variation in the area near the eyes. Zooming into the edited images of HFGI can observe unreal textures on the facial features. FeatureStyleEncoder broken the editability of $\mathcal{F}$ space is not able to achieve this semantic modification. StylePrompter instead is "smart" enough to capture the mood of smiling.}
  \Description{Additional editing results.}
  \label{editing_comparison_additional2}
\end{figure*}

\begin{figure*}[h]
  \centering
  \includegraphics[width=\textwidth]{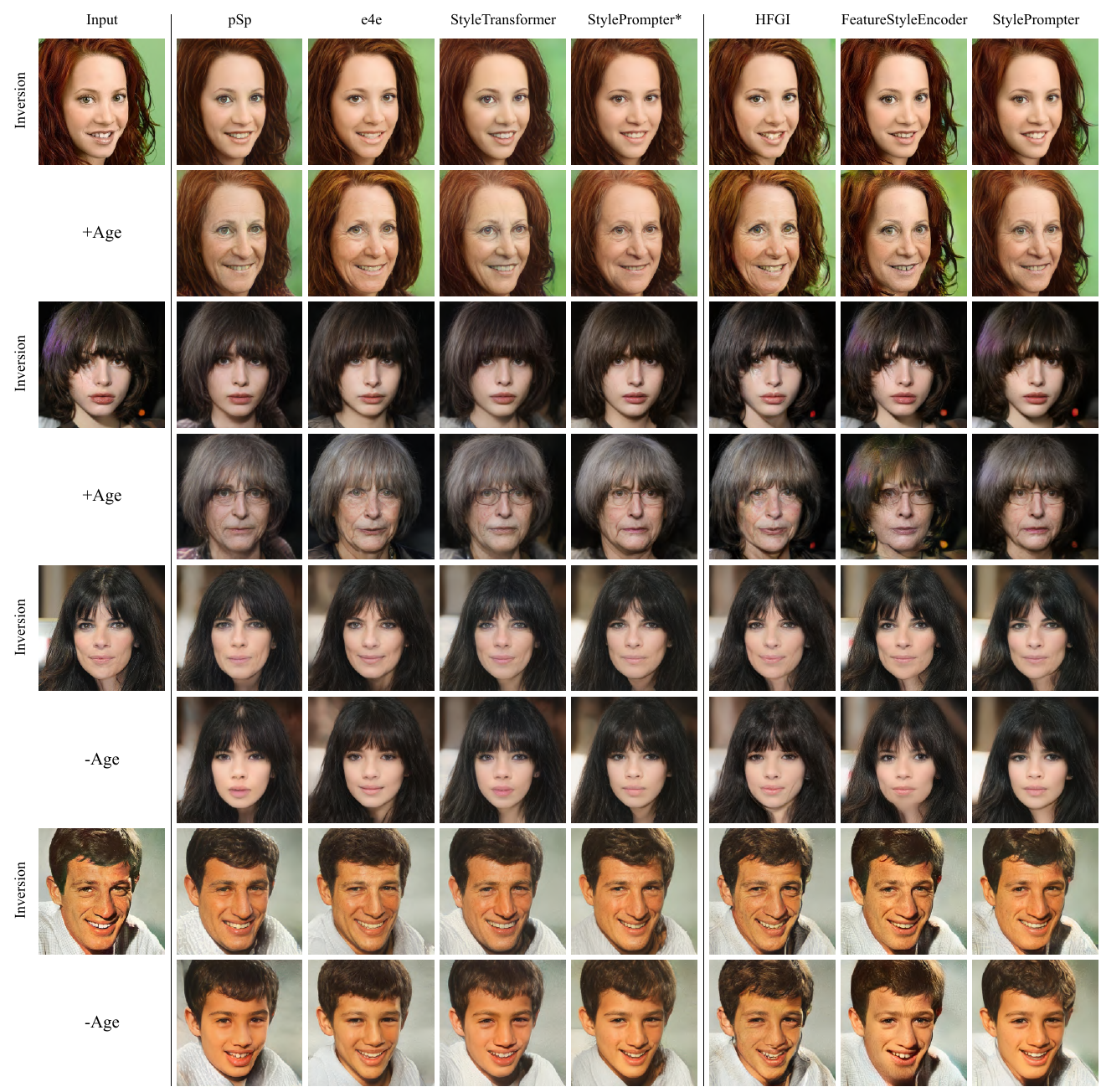}
  \caption{Editing comparison for $\mathcal{W^+}$-based methods (columns 2-5) and $\mathcal{F}$-involved methods (columns 6-8). Age manipulation is more challenging and semantic. HFGI and FeatureStyleEncoder both fail in producing real images, while our method takes full advantage of the modified style features and maintains the editability of $\mathcal{F}$ space. What's more, we carefully choose appropriate $\beta_1$ and $\beta_2$ to loosen the constraint of identity, thus achieving desirable edits.}
  \Description{Additional editing results.}
  \label{editing_comparison_additional3}
\end{figure*}

We provide more inversion comparisons in Figure \ref{inversion_results_simple} and \ref{inversion_results_hard}, editing comparisons in Figure \ref{editing_comparison_additional1}, \ref{editing_comparison_additional2} and \ref{editing_comparison_additional3}. We also compare the realism between $\mathcal{W^+}$-based methods and $\mathcal{F}$-involved methods in Figure \ref{realism_comparison}. To further understand the effects of different stages in the backbone, we conduct another ablation experiment in Figure \ref{ablation_stage}, which also demonstrates the efficiency of the multi-scale attention in SMART.

\subsection{Non-face Domain}
\begin{figure}
  \centering
  \includegraphics[width=\linewidth]{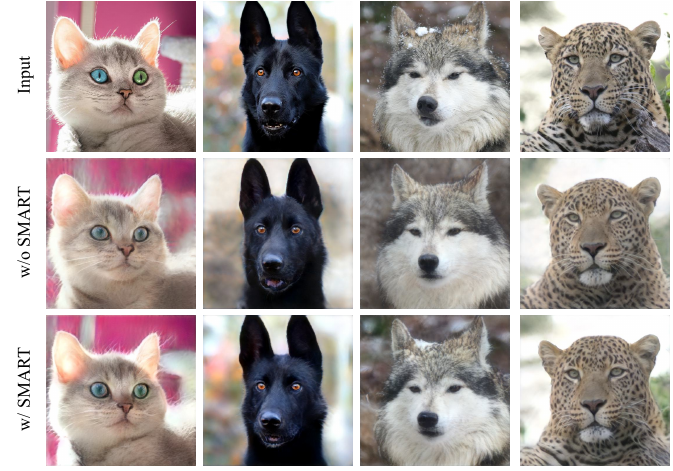}
  \caption{The inversion results for animal domain via StylePrompter. With SMART, inverted images can be more faithful to the input images.}
  \Description{SMART ablation in AFHQ.}
  \label{afhq_SMART_ablation}
\end{figure}

\begin{figure}[h]
  \centering
  \includegraphics[width=\linewidth]{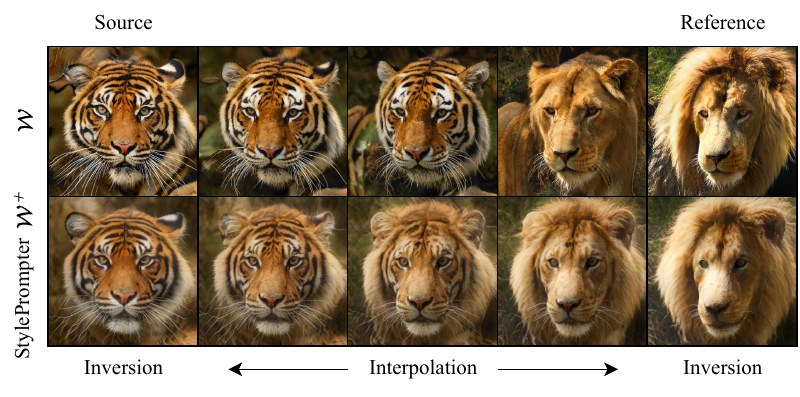}
  \caption{StylePrompter can also learn disentangled properties when transferring to the animal domain. Compare to the original $\mathcal{W}$ space, the interpolation results of StylePrompter baseline are smoother.}
  \Description{Interpolation comparison in AFHQ.}
  \label{afhq_compare_with_w}
\end{figure}

\begin{figure}[h]
  \centering
  \includegraphics[width=\linewidth]{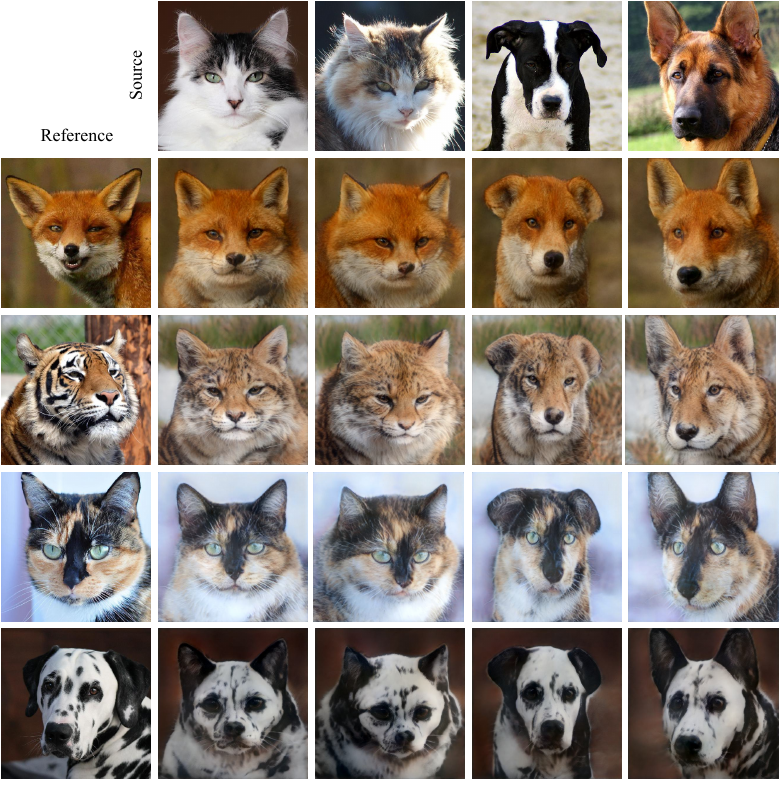}
  \caption{Style mixing in AFHQ animal domain. The source images are inverted via StylePrompter with SMART, then replace the source images' latent codes in layers 6-15 with that of the reference images.}
  \Description{F style mixing results in AFHQ.}
  \label{afhq_reference}
\end{figure}

\begin{figure*}[h]
  \centering
  \includegraphics[width=\textwidth]{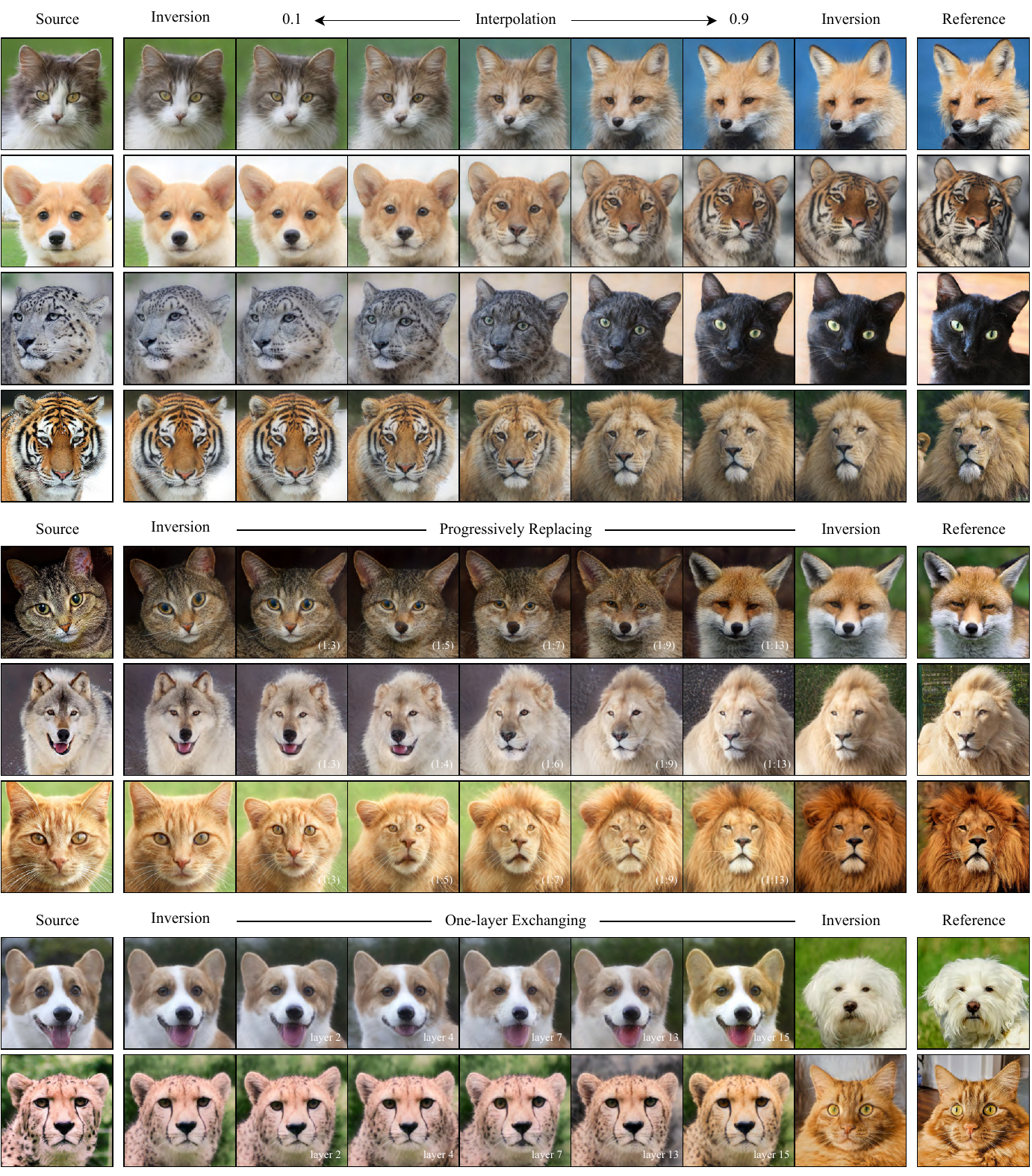}
  \caption{Three kinds of style mixing in the animal domain. 1-4 rows show the interpolation results, and 5-7 rows are the results of progressively replacing, and the last 2 rows show the results of one-layer exchanging. With more disentangled $\mathcal{W^+}$ space, the manipulations are of more semantic meanings.}
  \Description{Style mixing results in AFHQ domain.}
  \label{afhq_style_mixing}
\end{figure*}

We test StylePrompter on the animal domain using the generator pre-trained on AFHQ \cite{choi2020stargan} wild. To show the generalization ability of our method, we use full AFHQ including the cat, dog, and wild animal to train StylePrompter, as these animals have similar physical structures. We only exhibit the inversion and editing results based on the baseline and full model of StylePrompter, since some comparative methods have no available encoders in this domain. In Figure \ref{afhq_SMART_ablation}, we compare the inversion results with or without SMART. In Figure \ref{afhq_compare_with_w}, we show the disentanglement comparison between the baseline of StylePrompter in $\mathcal{W^+}$ space and the original $\mathcal{W}$ space using generated animal images. In Figure \ref{afhq_reference} and \ref{afhq_style_mixing}, we provide more style mixing results for the animal domain.

\begin{figure}[hbp]
  \centering
  \includegraphics[width=\linewidth]{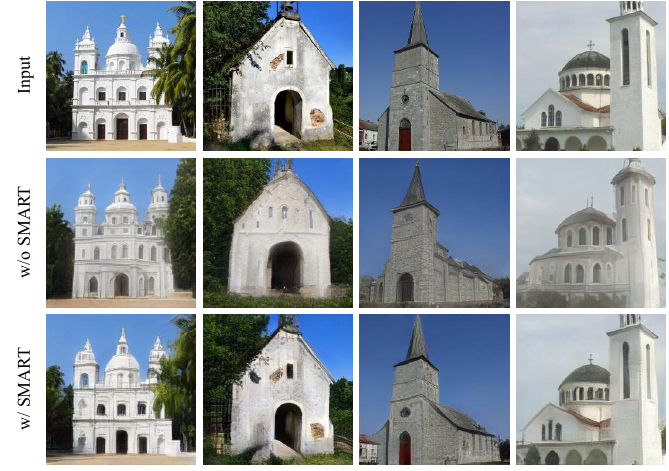}
  \caption{The inversion results for church domain via StylePrompter. With SMART, inverted images can be more faithful to the input images. The refinement of SMART in the church domain would be more challenging than the face domain, and the unreal patterns in the buildings can be easily recognized.}
  \Description{SMART ablation in church.}
  \label{church_SMART_ablation}
\end{figure}

We also provide additional results on the LSUN \cite{yu2015lsun} church dataset in Figure \ref{church_SMART_ablation}. This domain would be more challenging. The lower image resolution of the pre-trained church domain's StyleGAN generator and the greater diversity of this domain make the inversion more difficult, we suppose that is caused by inadequate style information for refinement when appending SMART.

\clearpage



\clearpage

\end{document}